\documentclass[10pt,twocolumn,letterpaper]{article}

\usepackage{cvpr}
\usepackage{times}
\usepackage{epsfig}
\usepackage{graphicx}
\usepackage{amsmath}
\usepackage{amssymb}
\usepackage{color}
\usepackage{subfig}
\usepackage[table]{xcolor}
\usepackage{algorithm2e}
\usepackage{rotating}
\usepackage{multirow}
\usepackage{arydshln}
\usepackage{makecell}

\usepackage[pagebackref=true,breaklinks=true,letterpaper=true,colorlinks,bookmarks=false]{hyperref}

\cvprfinalcopy 

\usepackage{titling}
\newcommand{\subtitle}[1]{%
  \posttitle{%
    \par\end{center}
    \begin{center}\large\texttt{\textbf{#1}}\end{center}
    \vskip0.5em}%
}

\ifcvprfinal\fi
\begin{document}

\title{Self-tuned Visual Subclass Learning with Shared Samples\\ An Incremental Approach\\}

\subtitle{Updated ICCV 2013 Submission}

\date{\vspace{-5ex}}

\author{Hossein Azizpour\\Royal Insitute of Technology(KTH)\\
 \tt\small azizpour@kth.se \and Stefan Carlsson\\
Royal Insitute of Technology(KTH)\\
\tt\small stefanc@csc.kth.se
}

\maketitle

\begin{abstract}
Computer vision tasks are traditionally defined and evaluated using semantic categories. However, it is known to the field that semantic classes do not necessarily correspond to a unique visual class (e.g. inside and outside of a car). Furthermore, many of the feasible learning techniques at hand  cannot model a visual class which appears consistent to the human eye. These problems have motivated the use of 1) Unsupervised or supervised clustering as a preprocessing step to identify the visual subclasses to be used in a mixture-of-experts learning regime. 2) Felzenszwalb et al. part model and other works model mixture assignment with latent variables which is optimized during learning 3) Highly non-linear classifiers which are inherently capable of modelling multi-modal input space but are inefficient at the test time.
In this work, we promote an incremental view over the recognition of semantic classes with varied appearances. We propose an optimization technique which incrementally finds maximal visual subclasses in a regularized risk minimization framework. Our proposed approach unifies the clustering and classification steps in a single algorithm. The importance of this approach is its compliance with the classification via the fact that it does not need to know about the number of clusters, the representation and similarity measures used in pre-processing clustering methods a priori. Following this approach we show both qualitatively and quantitatively significant results. We show that the visual subclasses demonstrate a long tail distribution. Finally, we show that state of the art object detection methods (e.g. DPM) are unable to use the tails of this distribution comprising 50\% of the training samples. In fact we show that DPM performance slightly increases on average by the removal of this half of the data.
\end{abstract}
\section{Introduction}
\begin{figure*}
\begin{center}
  \centerline{\includegraphics[clip=true, trim=0cm 0cm 4cm 0cm, width=1\linewidth]{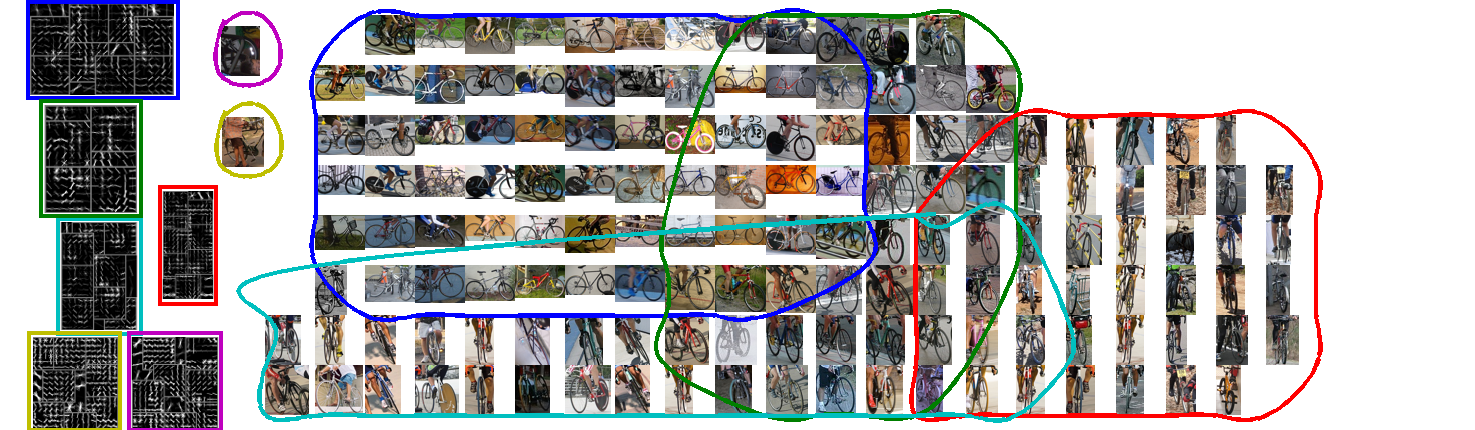}}
  \caption{A subset of visual subcategories found by our framework for the class of bicycle in the PASCAL VOC 2007 training set. The subclasses can share samples as long as they keep high accuracy on predicting them. The corresponding HOG filters for each subclass are marked by the same color of the border.\protect\footnotemark}
  \label{fig:teaser}
\end{center}

\end{figure*}
A limited set of visual classes such as frontal faces, pedestrians and
other objects detected in the context of urban traffic admit a rate of
detection that challenges that of human vision and make them
interesting for commercial
applications(\cite{OToole07,Dollar09}). This can be
attributed to the relatively limited degree of intra class variation
expressed by the images in these classes . On the other hand, multi
view detection of general object classes show a performance in the
range of around 20 to 65 \% in average precision (AP) (\cite{pascal-voc-2012}).  In recent
years, increase in the performance has come about by various
algorithms from year to year.  They are designed based on a limited
set of training data and are expected to capture
the class variation which is substantial due to different camera viewpoint, subordinate classes, etc. So, it is still an open question
whether this approach will eventually lead to the level of the performance of human visual system. The complexity of the classifiers to be learnt for these datasets is simply too large for the limited sets of data used to train them. The obvious remedy of trying to increase training data does not necessarily lead to better classifiers due to the emerging influence of outliers(\cite{ZhuBMVC12}).
To get around this we suggest an alternative way of designing visual classifiers. We consider the ultimate goal of a classifier is
to achieve performance at the level of human vision. However, instead of directly designing a classifier that works for a full training set, we examine subsets of the class for which classifiers can be
learnt with good performance. The objective of this design
is to increase the size of these subsets and possibly merge them while maintaining the good classification rate. The "size" has to relate in
some sense to the variation within the subclass. As these subclasses
grow and merge with each other it is required to increase the 
complexity of the classifiers applied to them. 
This approach should be contrasted with the prevailing methods of
designing classifiers for the full dataset that usually lead
to complex non-linear optimization problems. In order to handle these,
the optimization methods have to be regularized in order to avoid over
fitting. The proper information for regularization is simply not there at the start but has to be acquired gradually from the data itself.  We further claim that instead of directly competing in the performance of the final classification rate, competition should be in terms of size and complexity of subclasses that are properly classified.\\

\begin{figure*}
  \mbox{}\\
  \centering
  \includegraphics[clip=true, trim=0.5cm 6.4cm 2cm 6cm, width=0.22\textwidth]{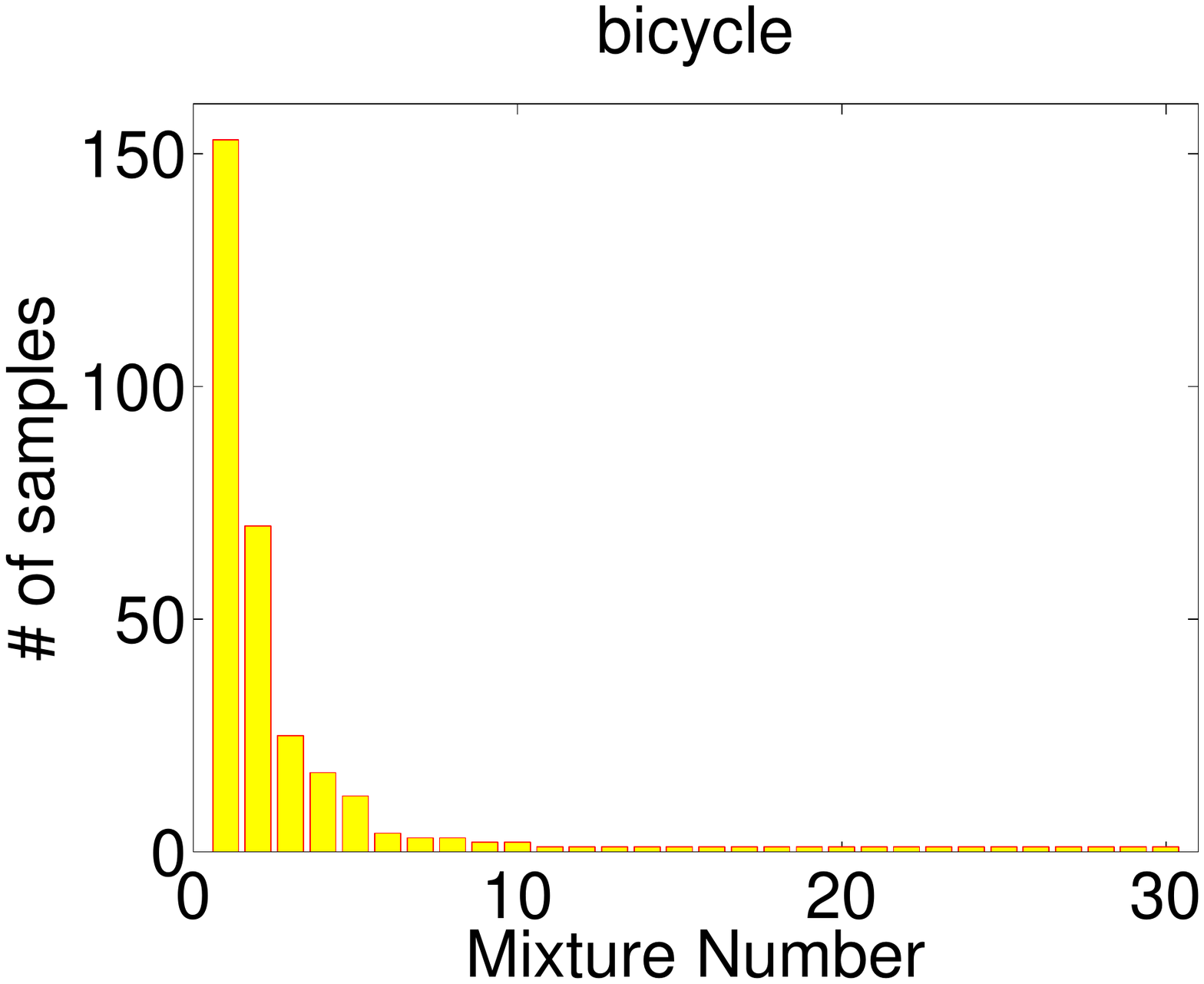}  
  \includegraphics[clip=true, trim=0.5cm 6.4cm 2cm 6cm, width=0.22\textwidth]{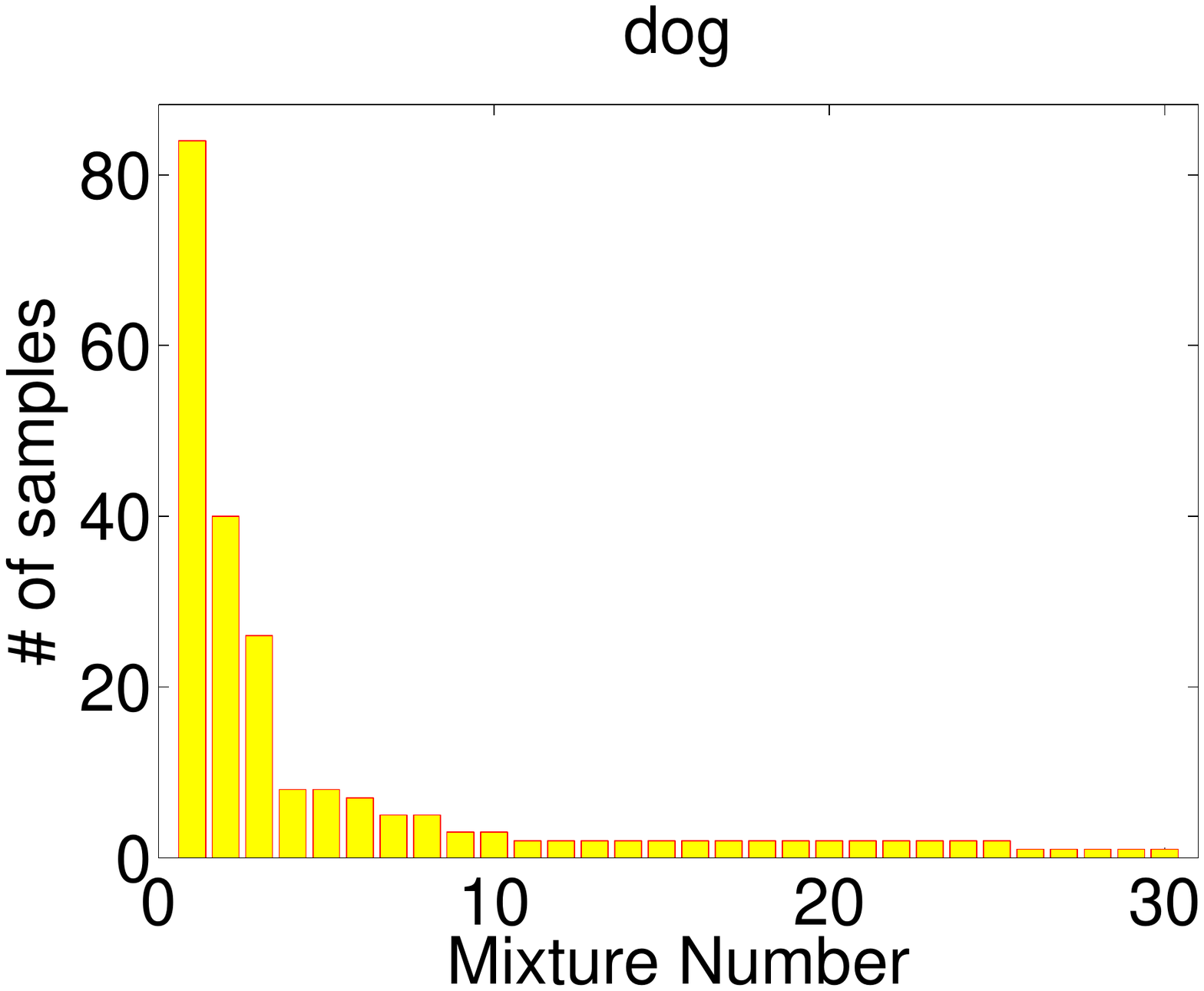}
  \includegraphics[clip=true, trim=0.5cm 6.4cm 2cm 6cm, width=0.22\textwidth]{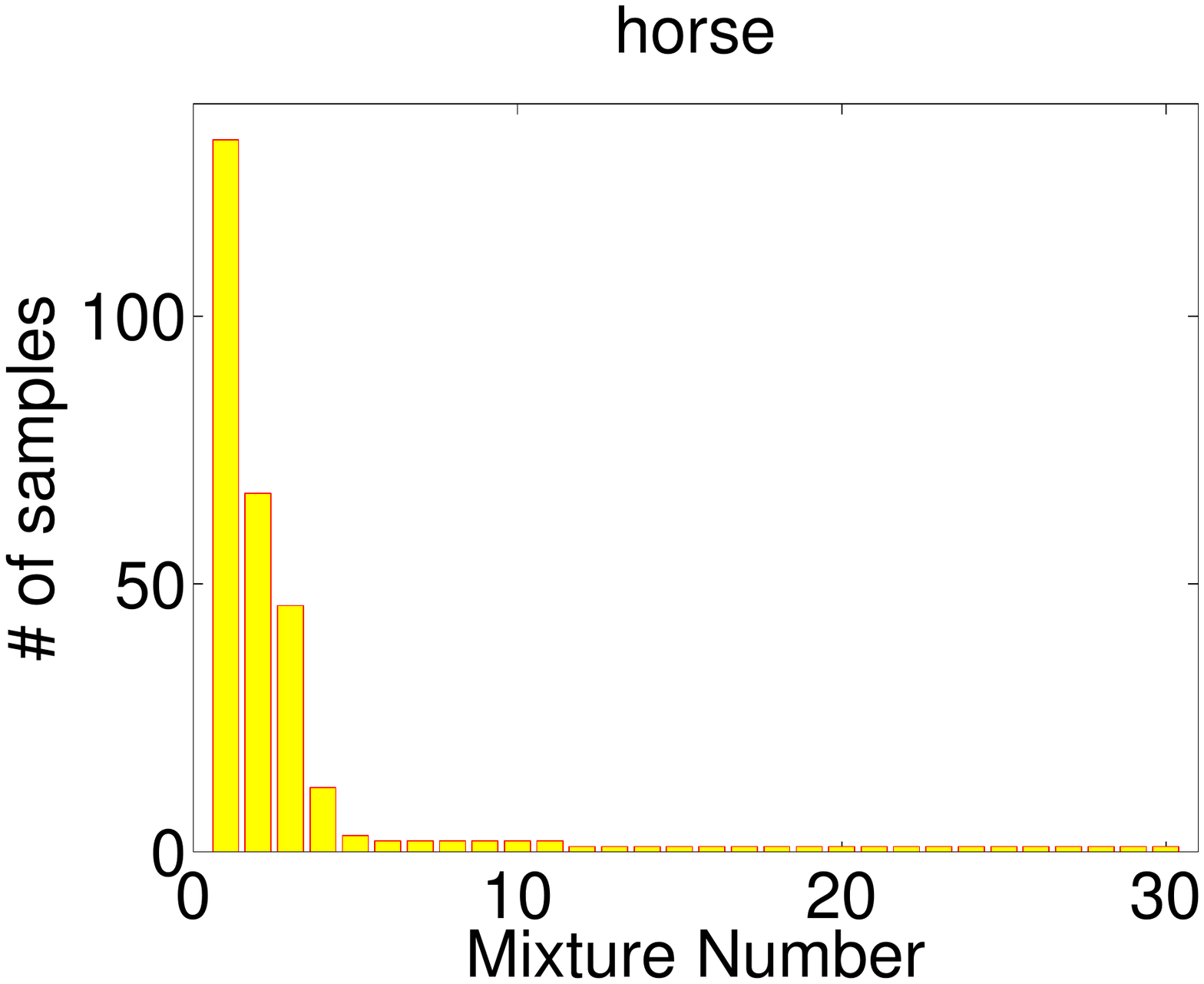}
  \includegraphics[clip=true, trim=0.5cm 6.4cm 2cm 6cm, width=0.22\textwidth]{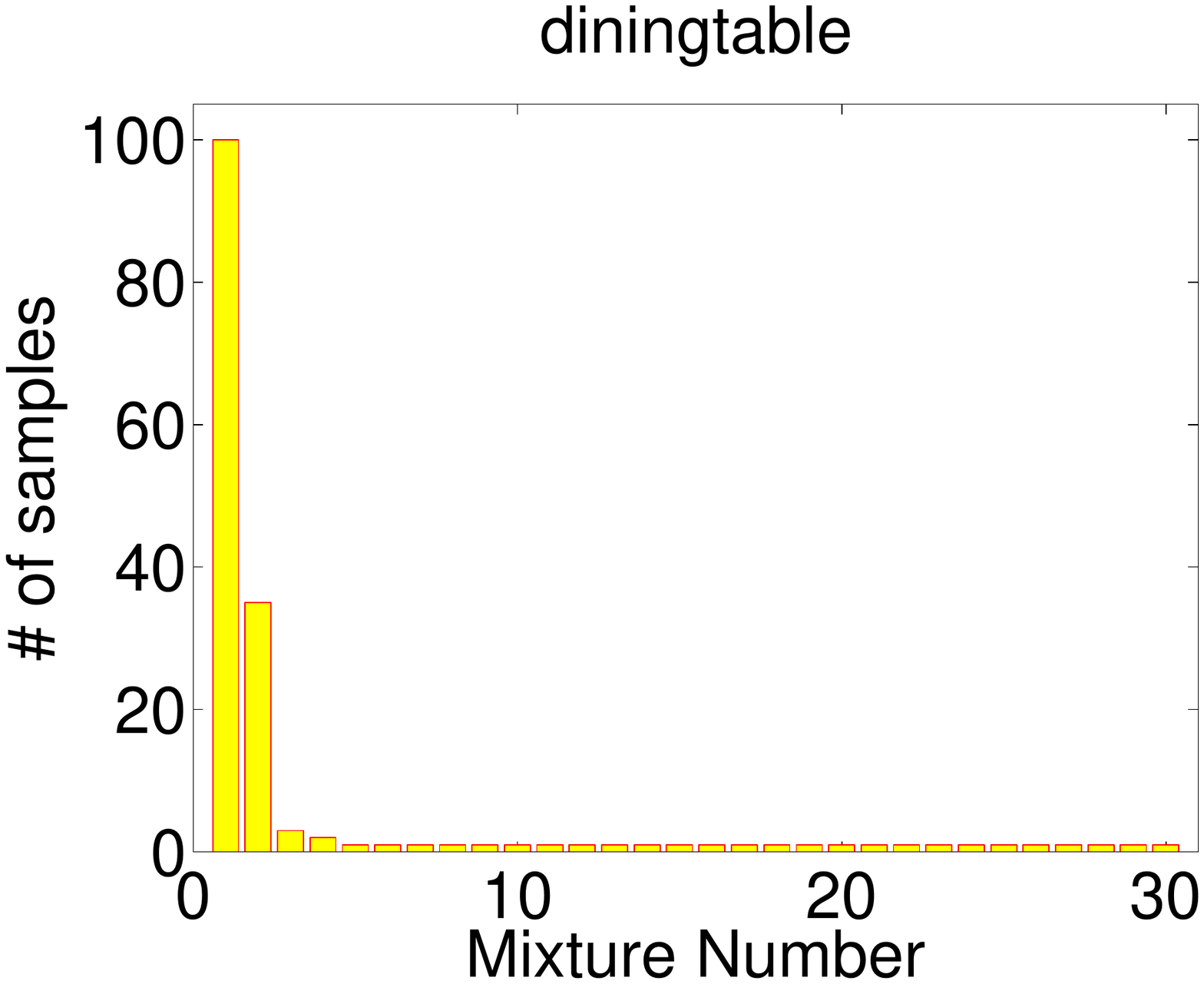}

  \caption{Distribution of the number of samples inside each visual subclass sorted in descending order for the first 30 components. A long tailed distribution is observed.}
  \label{fig:dist}
\end{figure*}

In the current work, we devise an incremental discriminative procedure which is able to simultaneously discover and learn such visual subclasses (Figure \ref{fig:teaser}). In particular, 
we formulate the learning problem as regularized risk minimization to find the largest subset which can be reliably detected. Taking the subset out we continue the same procedure 
on the remaining samples and so forth. Intuitively, this will leave us with a long-tailed distribution of subclasses (Figure \ref{fig:dist}). For the final experiments we specialize to the deformable part models (DPM) \cite{Felzenszwalb09} as our RRM framework.  \\This iterative procedure helps to keep outliers --that might harm the performance of subclasses-- away as isolated exemplars. 
This means that our work relates primarily to the so called mixture models of classification \cite{Felzenszwalb09} \cite{Yang13} \cite{Aghazadeh12}. 
This incremental approach, by finding "visual subclasses", allows understanding  of the problem at hand by analyzing the core large subsets \emph{and} isolated exemplars. It should be mentioned that by a "visual subclass" we mean a subset of our semantically defined class that can be reliably detected given a learning framework and feature representation and thus does not necessarily comply with visual subclasses to the human eye. Saying that, however, we can see in Figure (\ref{fig:teaser}) that the subclasses are still meaningful in appearance.
Finally, we posit that the recent success of Deep Learning methods trained on large outside-training datasets for visual recognition tasks \cite{Razavian14} is related to unification of learning representations that are invariant to such subclass variations and classification. The recognition problem is still unsolved, thus the advantage of our procedure to the deep learning methods is that we have a systematic analysis of where our recognition system fails which similarly highlights the importance of handling the tails of the distribution in deep learning frameworks.

\footnotetext{This figure is generated automatically using a subset of each subclass members, for larger pictures of all the classes please refer to the supplementary material.}

\subsection{Contributions and Outline}
The contributions of this work can be listed as follows,\\
\textbf{I)} A general incremental view on the recognition task is motivated: i.e. the identification, growth and merging of the \emph{maximal} subclasses which can be detected with \emph{high accuracy}.\\
\textbf{II)} We propose a simple yet effective discriminative procedure based on self-paced learning formulation for simultaneous discovery and learning of such subclasses (without the prior knowledge about the number of subclasses, the representation space or similarity measure suitable for clustering) which fits into the regularized risk minimization scenario. The formulation, seen as a data-driven clustering framework, is able to span the spectrum between the two extremes of a)one exemplar classifier per sample and b)a monolithic classifier for all samples.\\
\textbf{III)} We show the proposed self-paced learning is able to detect outliers (w.r.t. to the base classifier) and is equivalent to the ramp loss formulation of SVM.\\
\textbf{IV)} We allow sharing of the samples between the mixtures and we show that this helps the final classification by regularizing the decision boundaries where different mixture classifiers meet.\\
\textbf{V)} We show that the visual subclasses follows a long tail distribution.\\
\textbf{VI)} By an aggregation of the identified \emph{non-singular} subclasses (covering 15 to 80\% of the data) we show that state of the art methods (e.g. DPM) are unable to use, on average, ~50\% of data appearing at the tails of this distribution.\\
\textbf{VII)} We further investigate the statistics of the isolated exemplars in terms of different image properties which gives an insight for future works.
\\

In the next section we discuss the related works and highlight the differences. It is followed by the explanation of our method for learning visual subclasses. Then we report and discuss results from a set of experiments supporting the model aspects. Finally, we conclude the paper and mention interesting future directions.

\section{Related Works}
The idea of gradually designing learners is at the heart of the
methodology known as incremental learning \cite{Yeh08, Gomes08,
  Kembhavi09}. However unlike standard incremental learning we focus on the gradual construction of well classified subsets and
we assume that all training data is available at the
start, although incrementally adding training data could be part of
our scheme too.\\
Identification of the regions of the input space to be classified by each expert is the main question when using mixture of experts. Particularly for the task of classification on the more realistic vision datasets\cite{pascal-voc-2012}, different works exist. They are mainly comprised of partitioning the input space in some way \cite{Wang12} or hard disjoint partitioning of the input samples \cite{Collobert01} and then learning individual classifiers using those subsets. Some methods are provided with explicit assignment  cluster members. \cite{Gu10} uses viewpoint annotations. However, annotation of other sources of variation, e.g. subordinates is hard to define and costly to do. Other methods, starting with an initial clustering, employ an EM-like approach to update the clusters. This is done by letting different experts compete for samples \cite{Felzenszwalb09, Kim08, Song13}. These methods, however, are susceptible to initialization. \\ To improve the initialization, other works use a clustering methods on either the image patches itself \cite{Aghazadeh12} or on some available meta-data \cite{Bourdev09,Azizpour12,Gu12}. While they have indeed shown improvements over baselines, the number of clusters is unknown a priori. Furthermore, the clustering method might not be compatible with the final model since they are optimized for different objectives.\\
Our formulation is similar to that of self-paced learning \cite{Kumar10}, however, with a completely different perspective. Kumar et al. use the formulation to cope with the sensitiveness of optimizing latent variable models. And for this they gradually consider all the examples by starting from simpler examples. This is in contrast to our scheme. We want each subset to contain only the accurately predictable examples. So, conversely, we start by including more examples at the start and gradually converge to the largest core of the samples that can be reliably detected.\\
Our framework, can be seen as a discriminative clustering. Our approach, however, is different from other existing discriminative clustering methods \cite{Bach07,Gomes10,Hoai13} in the sense that it tries to best discriminate the positive and negative classes. This is unlike the conventional discrimination of the positive clusters that we actually avoid by sharing samples. \cite{Razavi12,Torsello08} also demonstrate the effectiveness of overlapping clusters. In addition our proposed clustering method does not require prior knowledge about the number of subclasses, the representation space or similarity measure suitable for clustering. \\
Finally, our framework is closely related to \cite{Doersch13,Doersch12} for discovery of midlevel features used in tasks other than detection. However, we see our proposed method as a more principled strategy for discovery of subclasses which can be used to group such mid-level representations as well.

\section{Methodology}
Here we explain the framework used to train our self-tuned mixture classifiers. First we recap the Regularized Risk Minimization problem (RRM) and  mixture of experts(MEX). Then we introduce a new formulation for learning each expert of MEX and discuss its properties. Finally, by injecting DPM as the individual RRM learner we show how we optimize the objective in an iterative manner.
 
\subsection{Mixture of Regularized Risk Minimizers}
Discriminative learning methods have been very successful in many applications of machine learning such as computer vision. This is particularly true for classification tasks when provided with a relatively large set of labelled training samples. In the context of \emph{discriminative} learning of binary classifiers, given a set of $N$ labelled training samples $D=\{\langle x_i,y_i\rangle\}_{i=1}^{N}, y_i\in\{-1,1\}$, we are usually faced with a RRM formulation of the form,

\small
\begin{align}
\label{eq_RRM}
&\min_{\mathbf{w}}J(\mathbf{w}):=\Omega(\mathbf{w})+\lambda R_{emp}(\mathbf{w}, D)\\
&R_{emp}(\mathbf{w},D)=\sum_{i=1}^{N}l_{i}^\mathbf{w}
\end{align}
\normalsize

where $l_{i}^{\mathbf{w}}$ is the "loss" of sample $i$. Solving an RRM involves a trade-off between the regularization term $\Omega$ and the risk term $R_{emp}$ which is controlled by the coefficient $\lambda$. There are different choices of a reguralizer e.g. L1, L2 and loss functions e.g. least squares or hinge (for Lasso or Support Vector Machines respectively). Depending on these choices there are well-established solvers in the literature.\\
However, it has been observed in several applications that the distribution of the training samples in the input space is multi-modal. This is because of the semantically defined labels, data incompleteness and poor representation (e.g.. not invariant enough to different conditions). Although kernel methods or similar approaches can partially overcome these problems, they are usually inefficient for large scale problems and parameters should be carefully tuned for them to avoid overfitting. This has caused the mixture-of-experts (MEX) and ensemble methods to become popular. MEX uses a combination of simpler learners (e.g. linear SVMs) to train a more complicated classifier able to cope better with the multi-modality of the input space. It should be noted that MEX is subtly different from other ensemble methods in that it divide the input space by training each expert on either a hard subset or non-uniform distribution of training data. MEX of $M$ learners in terms of RRMs is defined as follows,

\small
\begin{align}
&\min_{W}J(W):=\Omega(W)+\lambda R(W,D)\\
&R(W,D)=\sum_{i=1}^{N}g(l_{i}^{\mathbf{w}_1},l_{i}^{\mathbf{w}_2}
,...,l_{i}^{\mathbf{w}_M})\\
&W=\{\mathbf{w}_1,\mathbf{w}_2,...,\mathbf{w}_M\}
\end{align}
\normalsize

where $g(.)$ is an aggregation function which can be anything from a gating function or a minimum of the losses produced by different experts. The learning of MEX involves two main steps, 1) identification and Learning of the parameters of the individual experts, 2) learning the parameters of the aggregation function. Some methods combine these two steps though \cite{Felzenszwalb09}. Due to this design, the three major difficulties of training a MEX are first, how to identify the partitions of the input space to be learned by an expert, secondly MEX is very likely to have problems at the boundaries where expert learners meet (these boundary regions usually have low or zero sample support), and finally what is a good choice of an aggregation function $g$. In the next section we introduce a new incremental approach as an effort to address the first two problems. The third issue of choosing an aggregation function, however, remains open for future work.
\subsection{Self-tuned Expert Learning}
We approach the problem of visual subclass discovery in an incremental fashion. By that we mean we are interested in finding the \emph{largest} subset of our positive training data ($y=1$) which we can distinguish from the negative data with a high precision. Then taking that subset away we continue by finding the next largest subset and so on. This will help the classifiers to train on more well aligned samples due to the removal of outliers. This helps us to better understand the classification problem we face by looking at more isolated components.  We pose the problem of finding the largest subset with high detection accuracy at each step using the following RRM optimization,\footnote{For the simplicity of the notation, we assume samples are ordered by their labels thus in the set $D$ there are $N_p$ positive ($y_i=1$) samples indexed by $i=1:N_p$ and then $N_n$ negative ($y_i=-1$) samples indexed by $i=N_p+1:N_p+N_n$.}

\small
\begin{align}
\label{eq_selftuned}
\min_{\mathbf{w},\mathbf{v}}J'(\mathbf{w})&:=\Omega(\mathbf{w})+\lambda R'(\mathbf{w},\mathbf{v},D)\\
R'(\mathbf{w},\mathbf{v},D)&=\sum_{i=1}^{N_p}v_i(l_{i}^{\mathbf{w}}-1/K)+
\sum_{i=1}^{N_n}l_{i+N_p}^{\mathbf{w}}\\
&\mathbf{v}=\{v_1,v_2,...,v_{N_p}\}, v_i\in\{0,1\}\nonumber
\end{align}
\normalsize

where $\mathbf{v}$ is a binary indicator vector that denotes which samples are included in the current mixture. $K$ is a scalar which controls how predictable a sample should be to get in the subset. Large $K$ is a tighter constraint on the predictability of the samples to be included. This is because as $K$ gets larger the loss $l_i$ of sample $i$ should be lower to be included. This formulation is interestingly similar to that of Kumar et al's self-paced learning\cite{Kumar10}. As described in \cite{Kumar10} for a convex choice of $l$ and $\Omega$ the relaxed formulation (i.e. $v_i \in [ 0,1 ]$) is a biconvex optimization in $\mathbf{w}$ and $\mathbf{v}$. This means that when we fix either of the subset of variables $\mathbf{w}$ or $\mathbf{v}$ the problem becomes convex over the other subset. Biconvex problems are specially interesting since they have good approximate solvers \cite{Gorski07} as well as branch and bound global optimizers.\\
Note that by choosing large enough or small enough $K$ we span a spectrum where we have Exemplar learners\cite{Malisiewicz11} at one extreme and one single learner at the other respectively.
It should be further noted that the formulation of self-paced learning is equivalent to the ramp loss formulation of \cite{Collobert06}. Ramp loss was originally designed to reduce the number of support vectors by ignoring the outliers which is essentially what we want in this approach. Further investigation of this equivalence and its consequences appears at appendix A in Sec \ref{sec:appendixA}.\\
Decision at boundaries where two independently trained experts meet can become ambiguous. To deal with this issue, after discovering each subclass, we let it share the already selected samples of previously trained experts. In other words we fix the corresponding elements of $\mathbf{v}$ to $1$  for the discovered subset and optimize (\ref{eq_selftuned}) for the whole training set. This will allow for sharing of samples between the experts which  makes them become smoother at the boundaries and potentially fill the gaps in the input space. In the experiments we found this is a crucial step for the performance of each expert. \\
Finally, this framework can be seen as a way of \emph{discriminative} incremental clustering. In that sense this clustering can potentially perform superior to conventional pre-processing clustering methods used for MEX. This is due to the fact that it unifies the partitioning of the positive samples in a way to comply better with the final classifier. That is our procedure does not require the number of subclasses, a new representation space or similarity measure suitable for clustering. 
\subsection{Self-tuned DPM}
To evaluate the effectiveness of our method for visual recognition we test it for the task of object detection. Since deformable part models (DPM) of Felzenszwalb at. al.\cite{Felzenszwalb09} is the current state-of-the-art, we use it as our individual experts. The formulation by replacing RRM terms with those of DPM then becomes as follows,

\footnotesize
\begin{align}
\label{eq_selftuned_dpm}
&\min_{\mathbf{w},\mathbf{v}}J'(\mathbf{w},\mathbf{v}):=\|\mathbf{w}\|^2+\lambda(\sum_{i=1}^{N_p}v_i(\max(0,1-f_\mathbf{w}(x_i) )-1/K)\nonumber\\ 
&+\sum_{i=1}^{N_n}\max(0,1+f_\mathbf{w}(x_{i+N_p})))\\\nonumber
&\mathbf{v}=\{v_1,v_2,...,v_{N_p}\}, v_i\in\{0,1\}
\end{align}
\begin{align}
\label{eq_dpm_score}
&f_\mathbf{w}(x_i)=\max_{h_i}(\mathbf{w}^T\Phi(x_i,h_i))
\end{align}
\normalsize

where $\Phi(.,.)$ is the HOG feature vector extracted from image $x_i$ at the location indicated by part latent variables $h_i$. DPM uses latent variables for modelling part positions in the scoring function $f$; Note that the loss function of positives in Eq. \ref{eq_selftuned_dpm} is a concave function. Thus (\ref{eq_selftuned_dpm}) is not a biconvex optimization anymore. However, as illustrated in \cite{Felzenszwalb09} the problems become convex as soon as one fixes the latent variables for the positive samples. We use this in the next section to come up with an Alternative Convex Search (ACS) algorithm to solve the above optimization problem.

\subsection{Optimization}
Solution of biconvex optimization problems can be well approximated using Alternative Convex Search (ACS) approach \cite{Gorski07}. This has been confirmed for Kumar et al.'s self-paced learning formulation of the structured-SVM \cite{Kumar10}. Using ACS one should fix one of the two disjoint subset of variables and solve the induced convex problem for the other subset and vice versa, and alternate until no improvement is possible. On the other hand Latent SVM formulation of DPM can be solved by alternative search between fixing latent variables and optimizing for $\mathbf{w}$ in the induced objective function. When $\mathbf{w}$ is fixed, the optimal solution of $v$ and $f_{\mathbf{w}}$ can be found in a single step (see Alg. \ref{alg:ACS}).\\

		\begin{algorithm}
			\SetAlgoLined
			\KwData {$D,K,C_p,C_n$}
			\Return {$\mathbf{w}$ model and $\mathbf{v}$ indicator vector}\\
			$\mathbf{v}\gets\{1\}$\\
			\While{$\mathbf{v}\neq \mathbf{v}_{old}$}{
			\textbf{0}: $\mathbf{v}_{old} \gets \mathbf{v}$\\
			\textbf{1}: Solve for $\mathbf{w}$ the convex program (\ref{eq_selftuned}) with fixed $\mathbf{v}$\\
			\textbf{2}: \For{$i=1:N_p$}{
				\If{$1-\mathbf{w}^T\mathbf{x}_i<1/K$}{
				$v_i=0$}
				\Else{$v_i=1$} 
			}
			}
			        \caption{ACS for self-paced learning}
        \label{alg:ACS}
		\end{algorithm}

		\begin{algorithm}
			\SetAlgoLined
			\KwData {$D,K,C_p,C_n$}
			\Return {$W$ set of trained experts $V$ set of indicator vectors}\\
			\While{$D'\neq \emptyset$}{
			\textbf{0}: $D'\gets D$\\
			$\,\,\,W,V \gets \emptyset$\\
			\textbf{1}: Solve (\ref{eq_selftuned_dpm}) for $\mathbf{v}$ and $\mathbf{w}$ using alg.\ref{alg:ACS}\\
			\textbf{2}: $V \gets V\cup \mathbf{v}$\\
			\textbf{3}: $W \gets W\cup \mathbf{w}$\\
			\textbf{4}: $D' \gets D' \backslash \{i|v_i=1\}$\\
			}
        \caption{Self-paced incremental}
        \label{alg:learning}
		\end{algorithm}

In an incremental approach, as illustrated in Alg. \ref{alg:learning} we solve the equation (\ref{eq_selftuned_dpm}) initially with the whole training set with Alg. (\ref{alg:ACS}) and then taking the indicated samples by $\mathbf{v}$ away, we repeat the procedure until no sample remains. 
It should be noted since we have an object detection scenario, we have huge number of possible negative windows from the negative images. Therefore, we adopt the hard negative mining approach of \cite{Felzenszwalb09} at step 4 of the algorithm (\ref{alg:ACS}).
\begin{figure*}[t]
  \mbox{}\\
  \centering
  \includegraphics[clip=true, trim=1cm 6.5cm 1cm 6.5cm, width=0.18\textwidth]{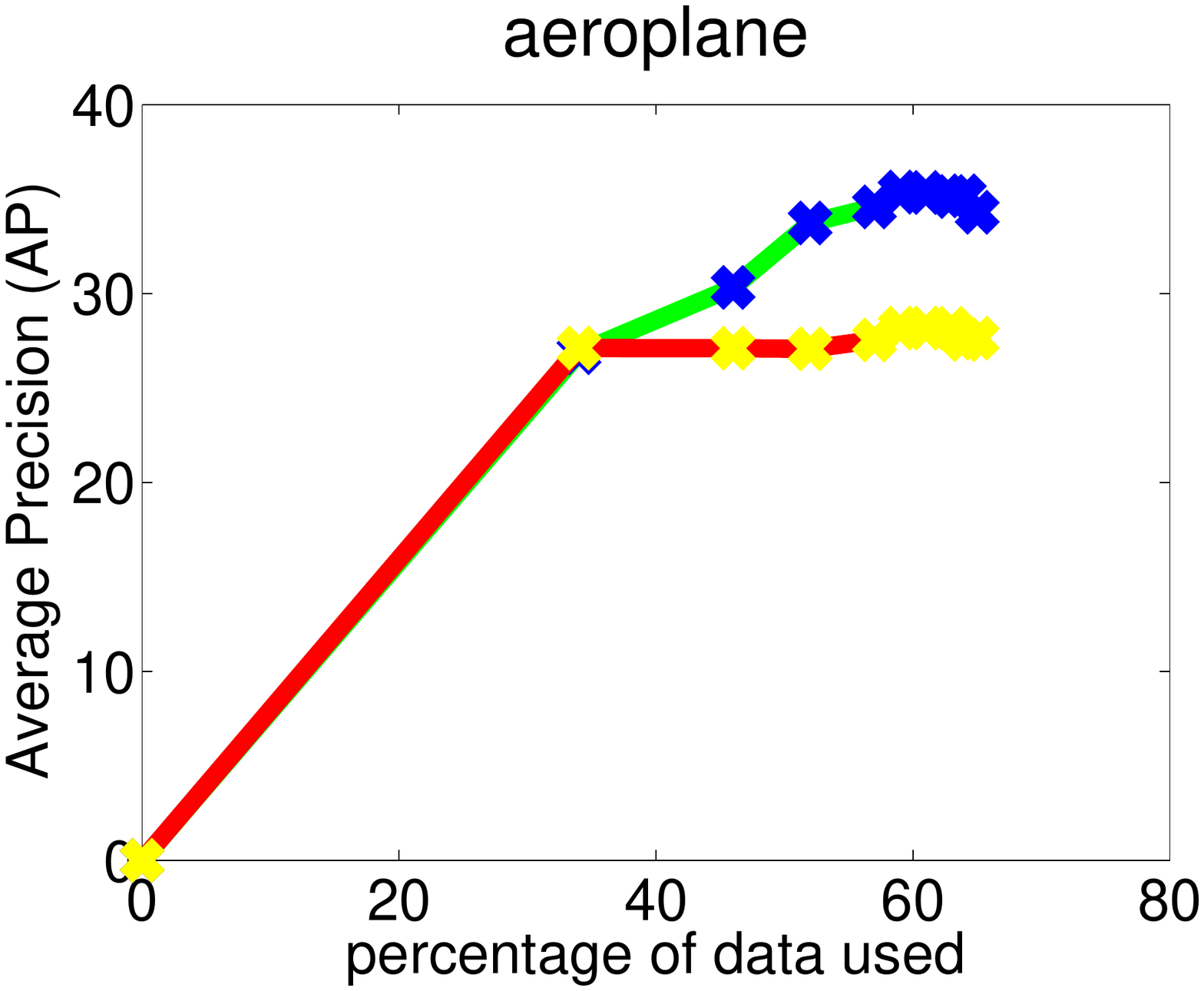}  
  \includegraphics[clip=true, trim=1cm 6.5cm 1cm 6.5cm, width=0.18\textwidth]{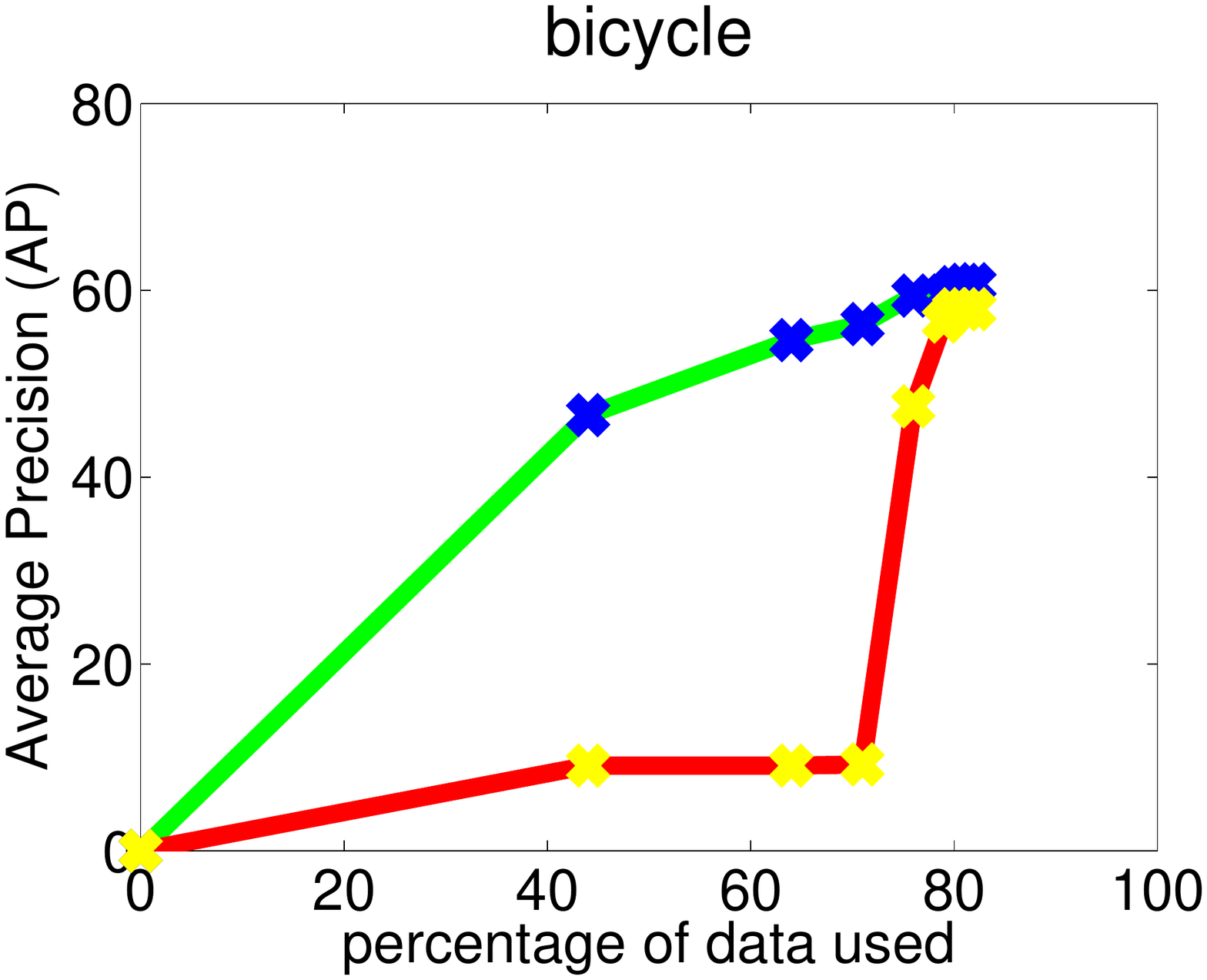}
  \includegraphics[clip=true, trim=1cm 6.5cm 1cm 6.5cm, width=0.18\textwidth]{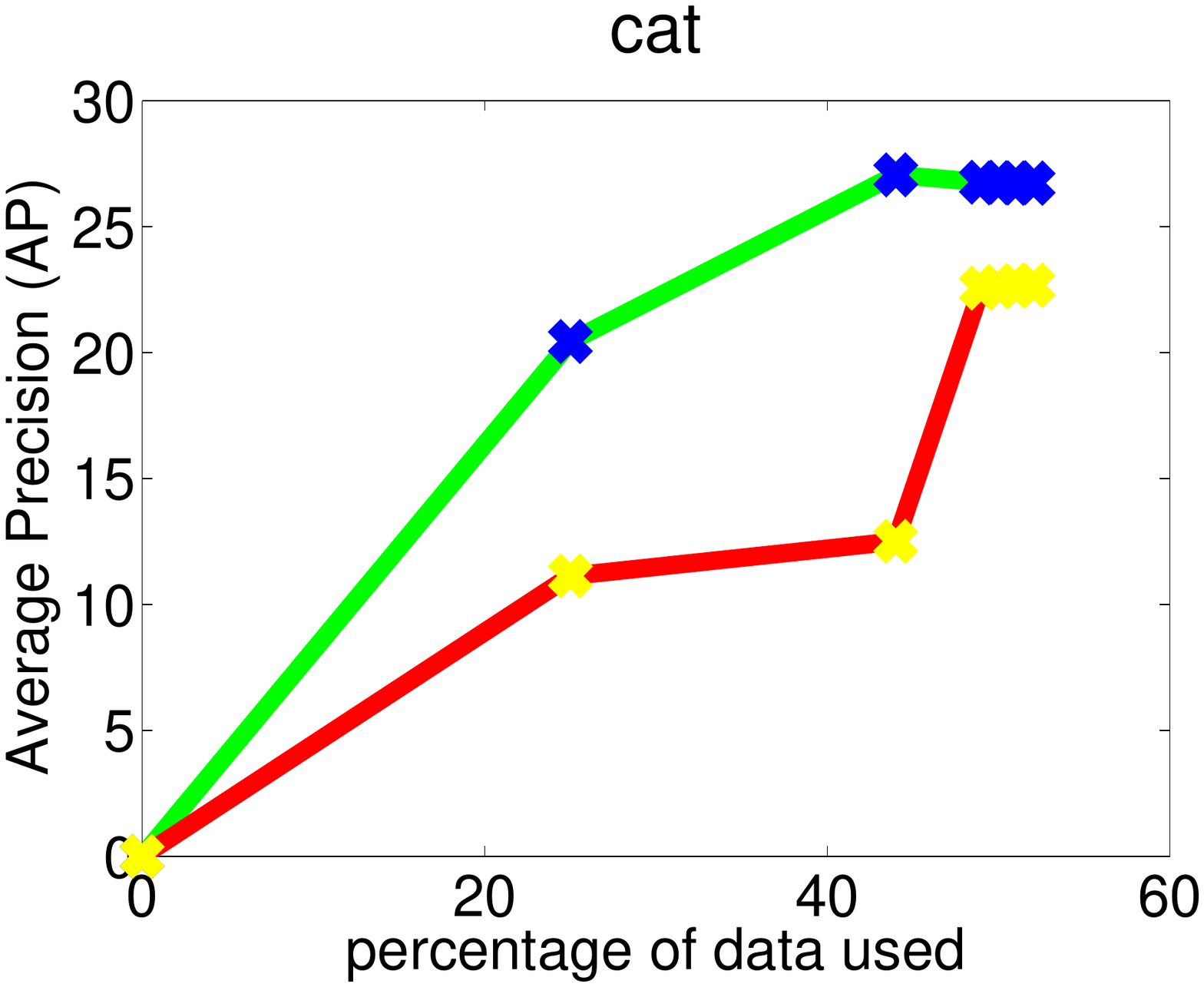}
  \includegraphics[clip=true, trim=1cm 6.5cm 1cm 6.5cm, width=0.18\textwidth]{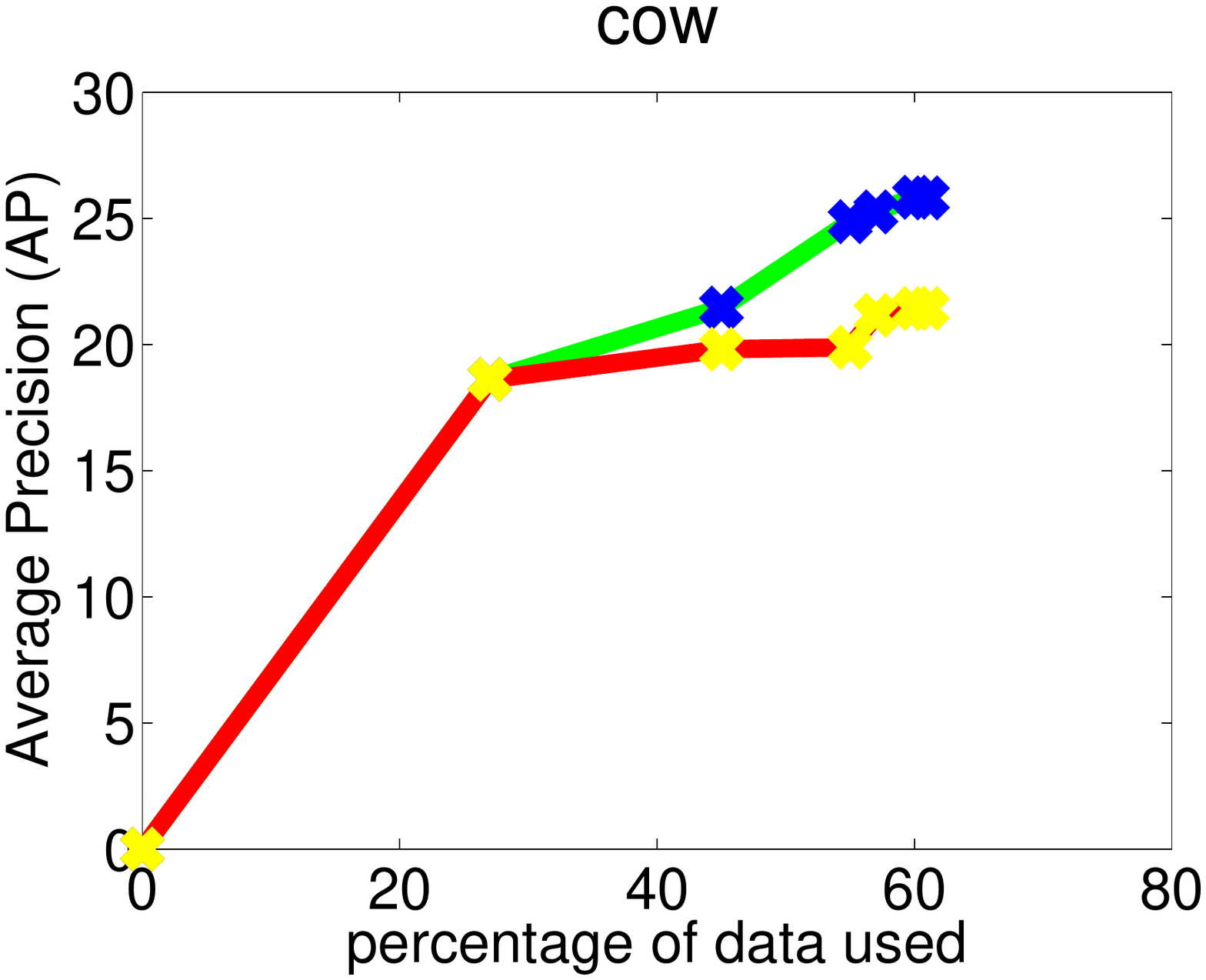}
  \includegraphics[clip=true, trim=1cm 6.5cm 1cm 6.5cm, width=0.18\textwidth]{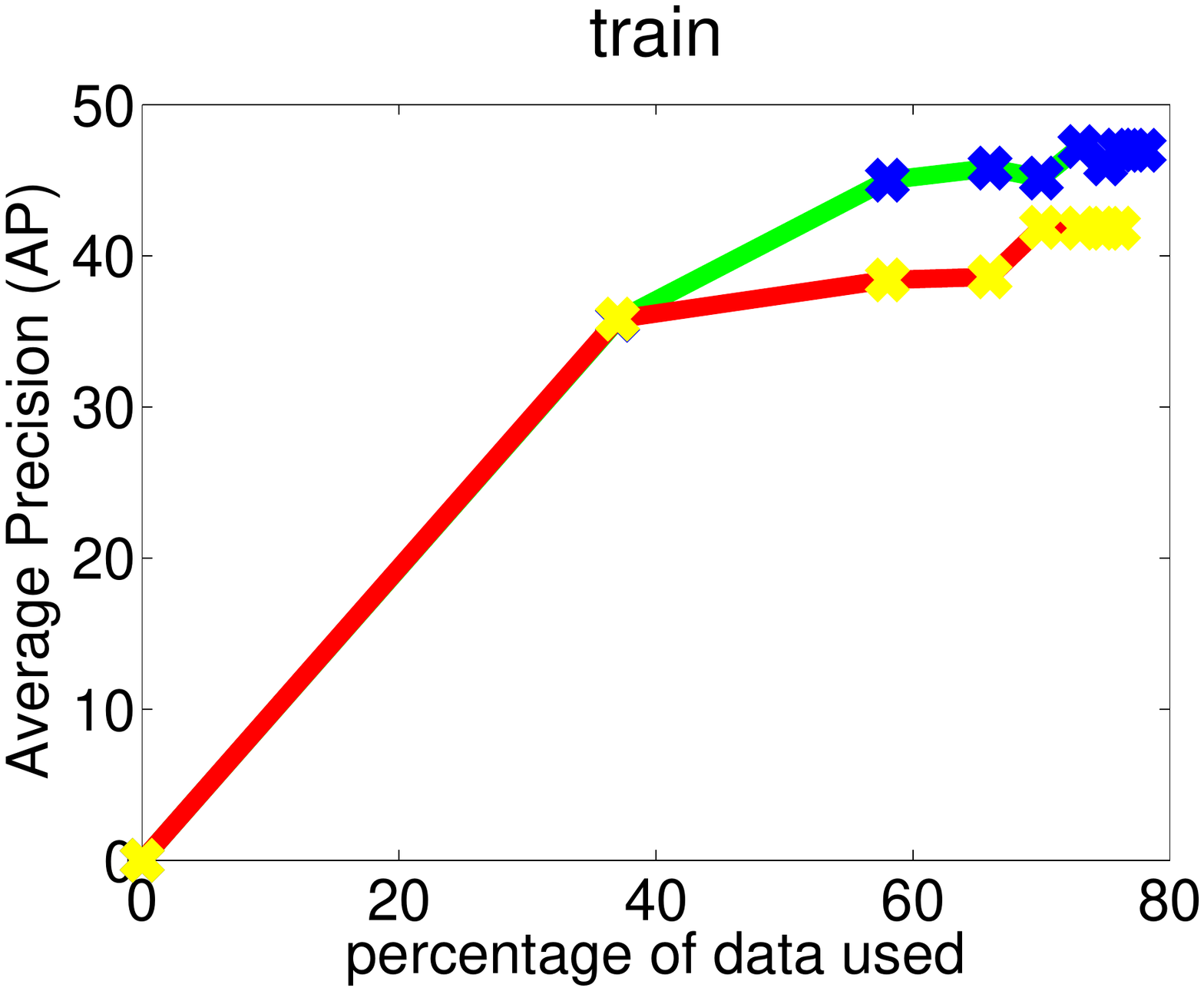}
  \caption{The evoultion of the AP when adding more experts for our full system (green -light gray- with blue crosses) compared with the case without sharing samples (red -dark gray- with yellow crosses) on 5 classes of VOC07.}
  \label{fig:APrevNog}
\end{figure*}
\section{Experiments}
We have chosen the task of Object Detection for evaluation of our method. The challenging dataset of PASCAL VOC 2007 is used as the testbed. In the following sections, we give the details of our implementation and then we continue by presenting the final detection performance of our system as well as showing quantitative and qualitative results illustrating different aspects of our framework. 
\subsection{Implementation Details}
We use the release 5 of the DPM \cite{voc-release5} as our base code without post processing of context re-scoring. We cross-validated a few choices of the parameters for the classes of "horse" and "car" on the validation set.\\ We set $K$ so that it contains all the samples on the positive half space defined by the decision boundary $\mathbf{w}$ ($K=0.7$). Furthermore, we found that starting with a lower $K$ in ACS (Alg. \ref{alg:ACS}) and then gradually increasing it to the final value will produce better results. It is initialized to a value that roughly includes the samples inside the SVM margin and on the positive half space($0.55$). Then $K$ is increased by a factor of $1.05$ to reach its final value.\\ This helps the optimization to better find the core subset at each iteration by taking relatively more samples in the beginning and then decreasing them if necessary. \\This approach is in contrast to \cite{Kumar10} where they start with higher value of $K$ and decrease it gradually. The aim there, is to consider all the samples eventually. They want to include cleaner sample first for the non-convex optimization of latent variables and then add the rest. We also tried to decrease the $K$ from a higher value and found that increasing $K$ gradually is a better choice for our purpose. The same gradual decrease in the $\lambda$, when the subset gets larger, is applied which in practice helped the convergence of the alternative search.\\
We do not have any particular aggregation policy to measure the final detection performance of the set of expert classifiers $W$. All the detections from all the experts are simply put together into the same pool (as if they were produced by a single classifier) and then a non-maximum suppression based on PASCAL VOC overlap criterion is applied.
\subsection{Object Detection on PASCAL VOC 2007}
To evaluate the final object detection performance, in favour of computation, we use a subset of trained experts. We only consider subsets that have at least 5 samples. This leaves us with less than 10 experts for each class. Table (\ref{table:voc07}) shows the final Average Precision(AP) on the 20 classes of VOC07 and it is compared with many of the existing state-of-the-art methods. Our method uses half of the data in average and yet it performs comparably to all the state-of-the-art methods. We can see that there is a marginal improvement in terms of mean AP over all classes. It is worth of mentioning that \cite{Chen10}, \cite{Vedaldi09} and \cite{Gu12} use multiple features (in addition to HOG). \cite{Gu12} uses meta-data for clustering. Finally, we plot the evolution of the AP as we add more experts in Figure (\ref{fig:APrev}). Observe that most of the performance is achieved at the very early steps. The most interesting future work, indicated by these results, is how to benefit from the rest of the isolated samples to push for better performance.

\begin{table*}[!t]
\scriptsize
\centering
\tabcolsep=0.15cm
\begin{tabular}{ l c c c c c c c c c c c c c c c c c c c c c }
     \Xhline{5\arrayrulewidth}
     &aero&bike&bird&boat&bottle&bus&car&cat&chair&cow&table&
     dog&horse&mbike&person&plant&sheep&sofa&train&tv&mAP \\ 
     \rowcolor[gray]{0.8}
    E-SVMs\cite{Malisiewicz11}&20.8&48.0&7.7&14.3&13.1&39.7&41.1&5.2&11.6 &18.6&11.1&3.1&44.7&39.4&16.9&11.2&22.6&17.0&36.9&30.0&22.7\\ 
    MCL\cite{Aghazadeh12}&29.0&50.1&9.6&15.0&18.9&41.1&49.7&10.3&16.0 &21.0&17.0&10.3&50.0&39.6&33.0&9.0&19.8&22.0&38.2&34.3&26.7\\ 
     \rowcolor[gray]{0.8}
    \cite{Gu12}&33.4&37.0&15.0&15.0&22.6&43.1&49.3&32.8&11.5&\textbf{35.8}&17.8
    &16.3&43.6&38.2&29.8&11.6&\textbf{33.3}&23.5&30.2&39.6&29.0\\
    AMH\cite{Chen10}&34.8&54.4&\textbf{15.5}&14.6&24.4&50.9&54.0&\textbf{33.5}&20.6&22.8
    &34.4&\textbf{24.1}&55.6&47.3&34.9&\textbf{18.1}&20.2&30.3&41.3&43.3&33.8\\
     \rowcolor[gray]{0.8}
    MKL\cite{Vedaldi09}&\textbf{37.6}&47.8&15.3&15.3&21.9&50.7&50.6&30.0&17.3
    &33.0&22.5&21.5&51.2&45.5&23.3&12.4&23.9&28.5&45.3&\textbf{48.5}&32.1\\
    DPM v5\cite{voc-release5}&33.2&60.3&10.2&\textbf{16.1}&\textbf{27.3}&54.3&\textbf{58.2}&23.0&20.0&24.1&26.7
    &12.7&\textbf{58.1}&\textbf{48.2}&\textbf{43.2}&12.0&21.1&\textbf{36.1}&46.0&43.5&33.7\\ 
     \rowcolor[gray]{0.8}
    Ours &34.3&\textbf{60.7}&10.3&13.7&21.8&\textbf{57.5}&55.5&26.7&\textbf{22.0}&25.8&\textbf{38.3}&
     14.3&57.0&47.7&43.0&9.0&19.8&34.4&\textbf{47.0}&43.7&\textbf{34.2}\\
    \hline
    \% of Data&65\%&82\%&28\%&25\%&45\%&70\%&73\%&52\%&43\%&61\%&66\%&38\%&74\%&69\%&63\%&14\%&36\%&50\%&78\%&71\%&55\%\\
	\hline    
\end{tabular}
\caption{Results of our method for object detection in VOC07 compared to other state-of-the-art baselines. The last row indicates the fraction of the training data used for each class to achieve the reported AP. It should be noted that \cite{Chen10},\cite{Vedaldi09} and \cite{Gu12} use additional features to HOG. \cite{Gu12} uses meta-data for clustering.
}
\label{table:voc07}
\end{table*}

\subsection{Aggregation}
\begin{figure*}
  \mbox{}\\
  \centering
  \includegraphics[clip=true, trim=1cm 6cm 1cm 8cm, width=0.21\textwidth]{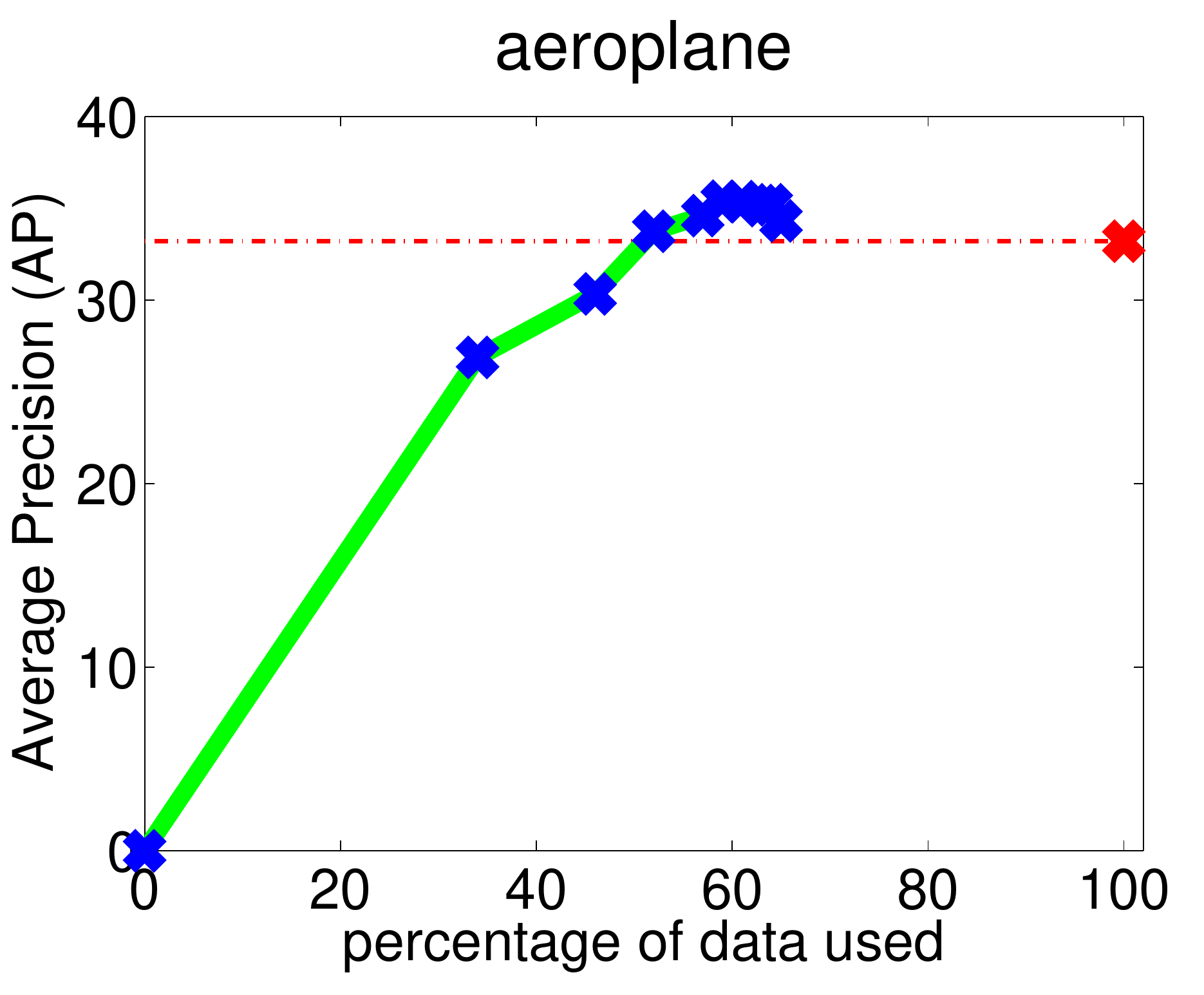}  
  \includegraphics[clip=true, trim=1cm 6cm 1cm 8cm, width=0.21\textwidth]{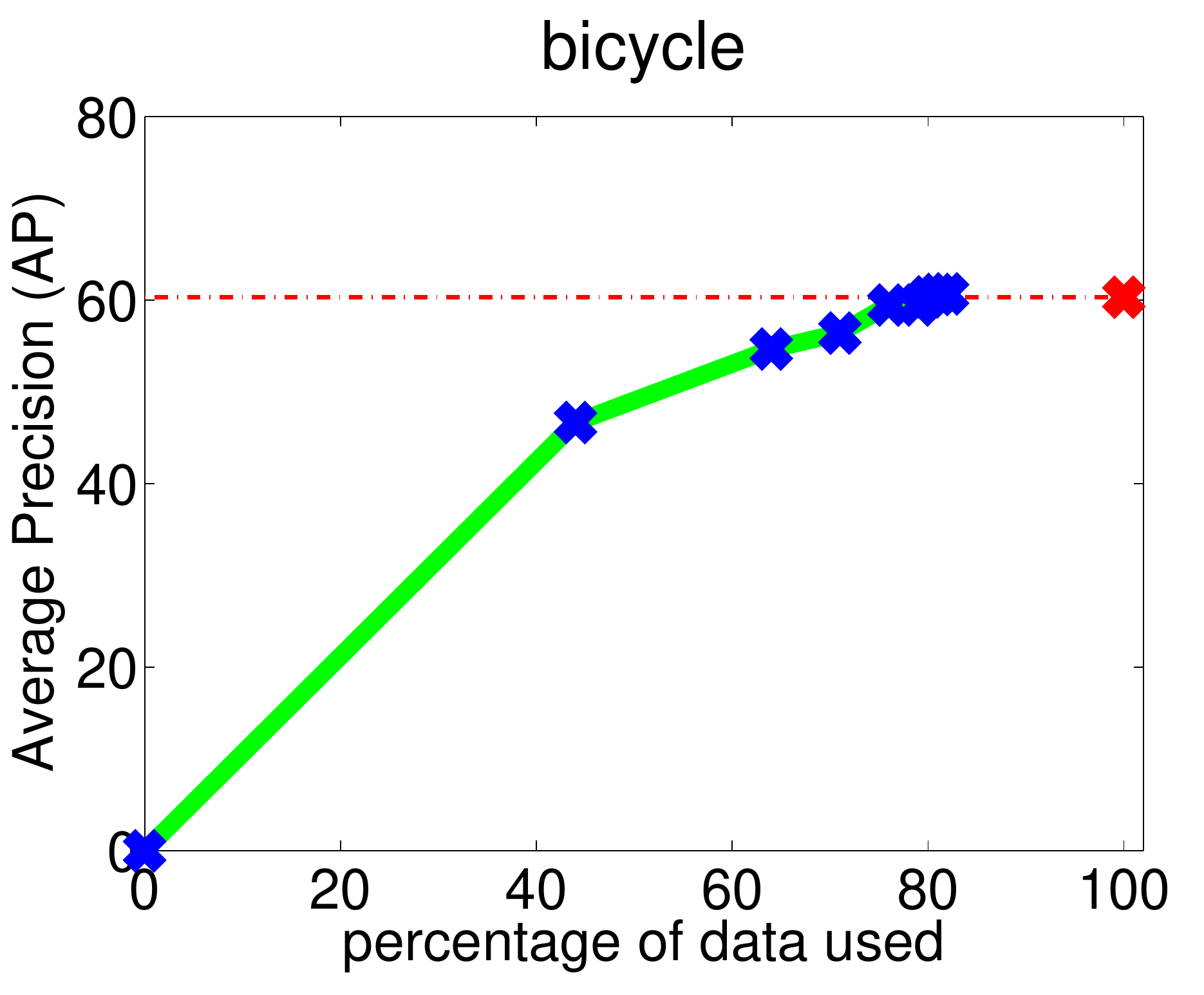}
  \includegraphics[clip=true, trim=1cm 6cm 1cm 8cm, width=0.21\textwidth]{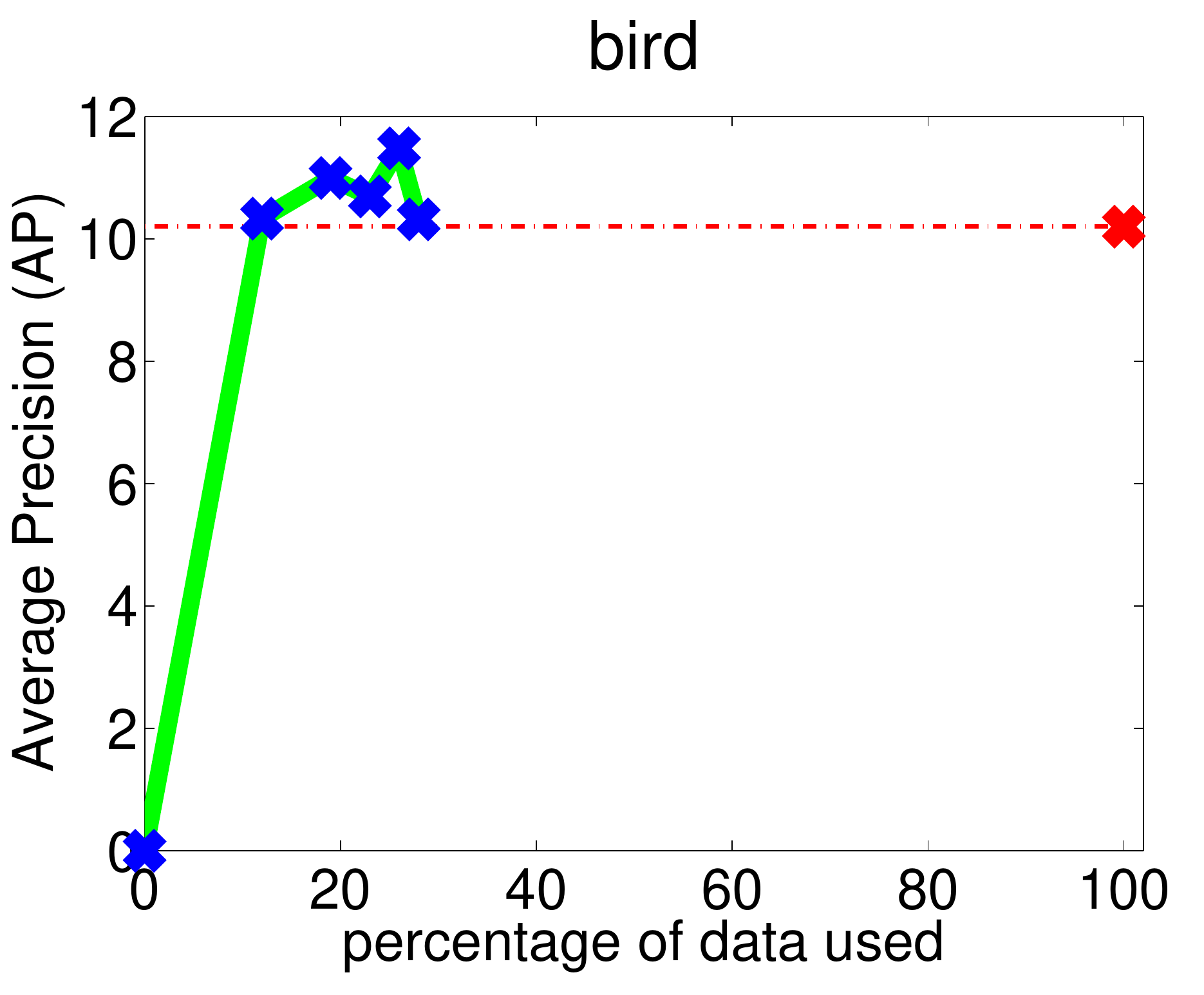}
  \includegraphics[clip=true, trim=1cm 6cm 1cm 8cm, width=0.21\textwidth]{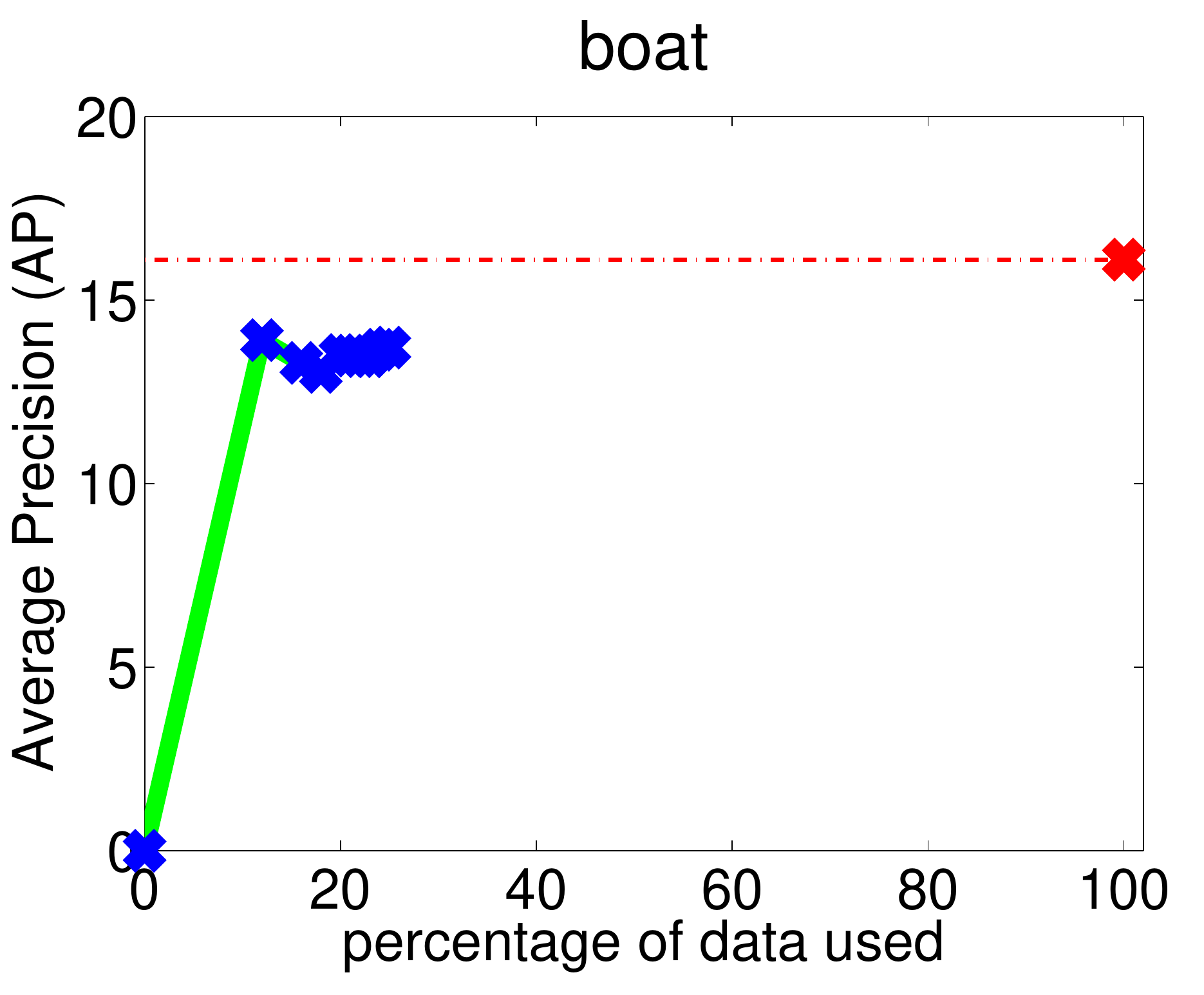}\\
  \includegraphics[clip=true, trim=1cm 6cm 1cm 8cm, width=0.21\textwidth]{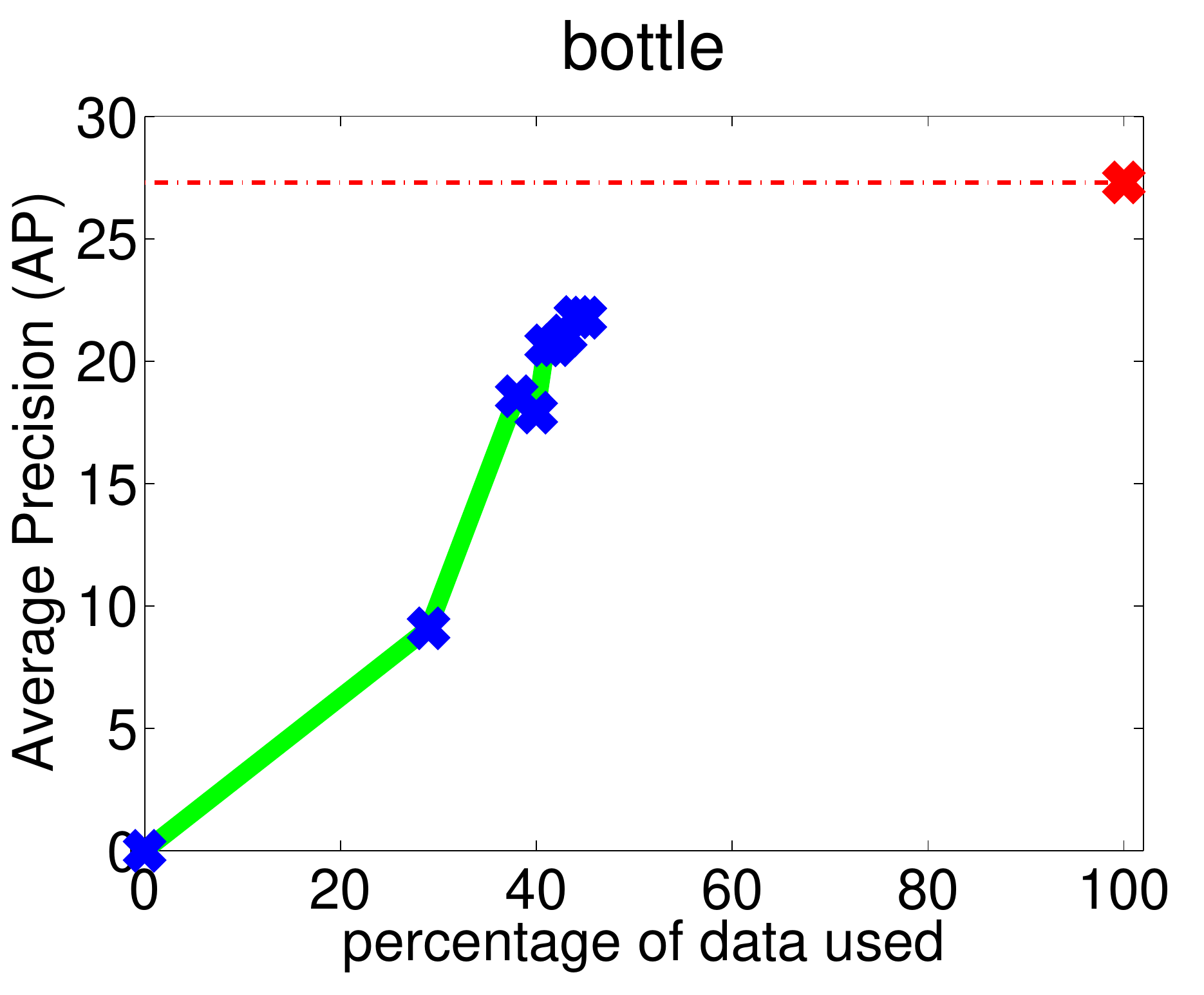}
  \includegraphics[clip=true, trim=1cm 6cm 1cm 8cm, width=0.21\textwidth]{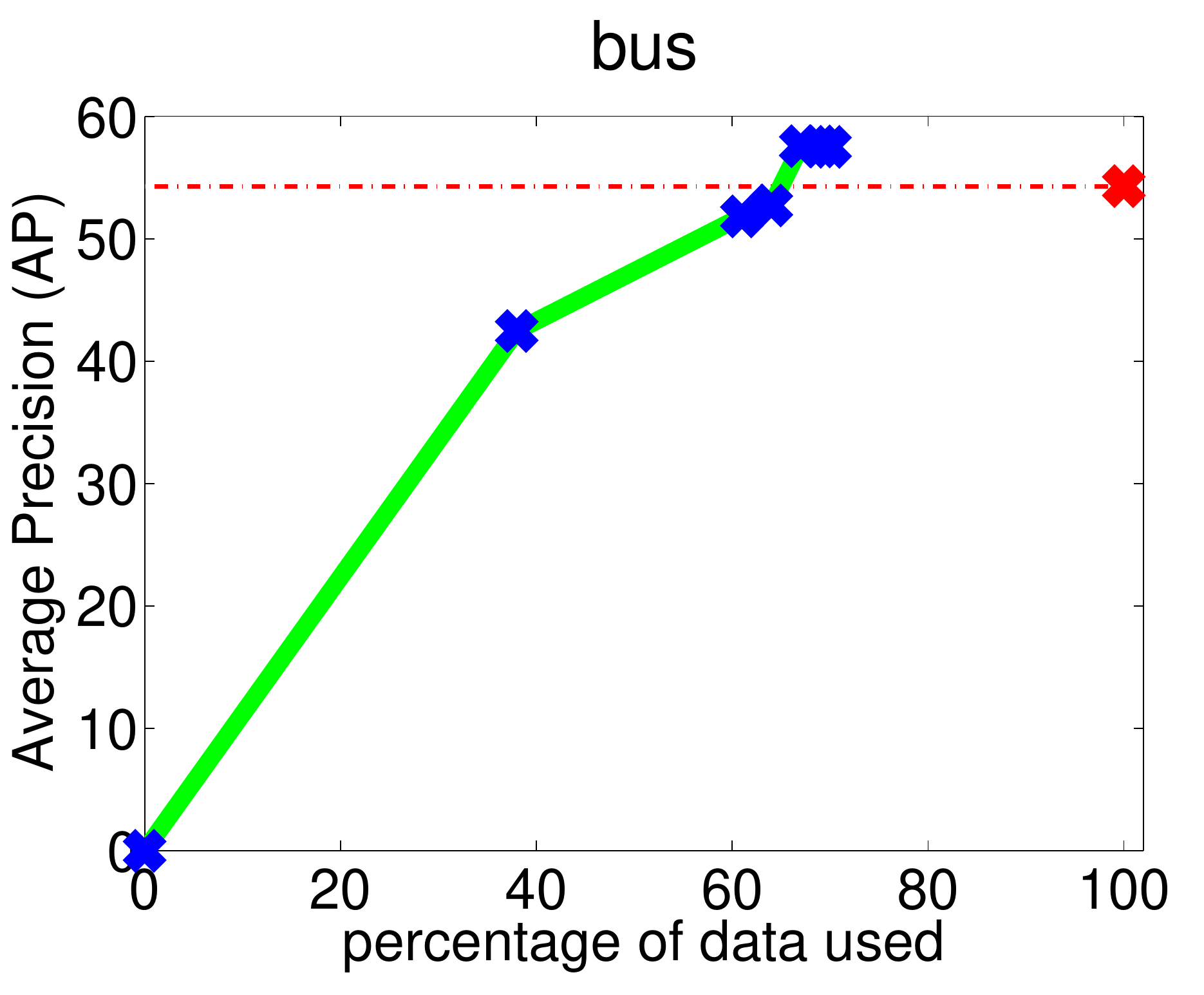}
  \includegraphics[clip=true, trim=1cm 6cm 1cm 8cm, width=0.21\textwidth]{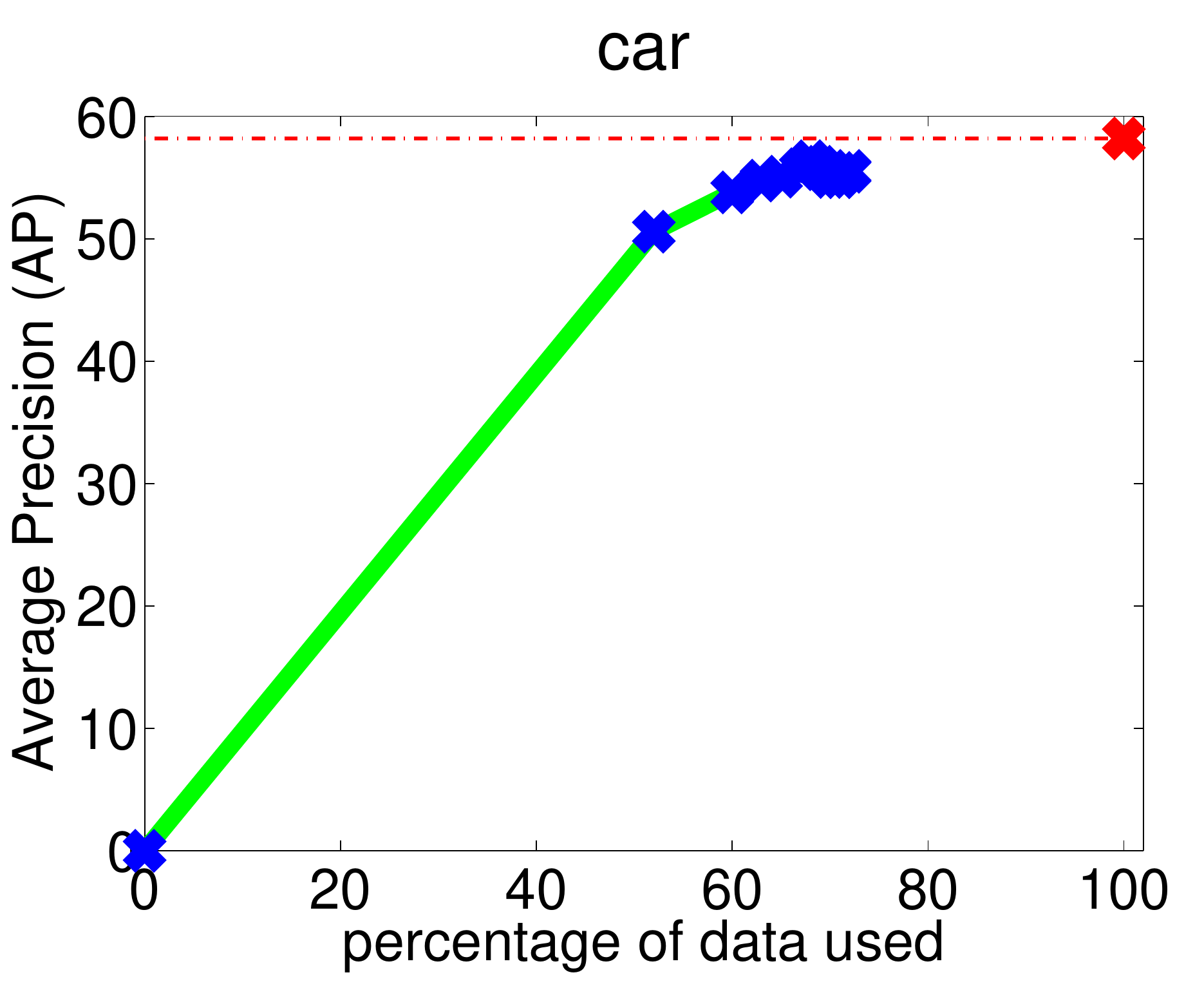}
  \includegraphics[clip=true, trim=1cm 6cm 1cm 8cm, width=0.21\textwidth]{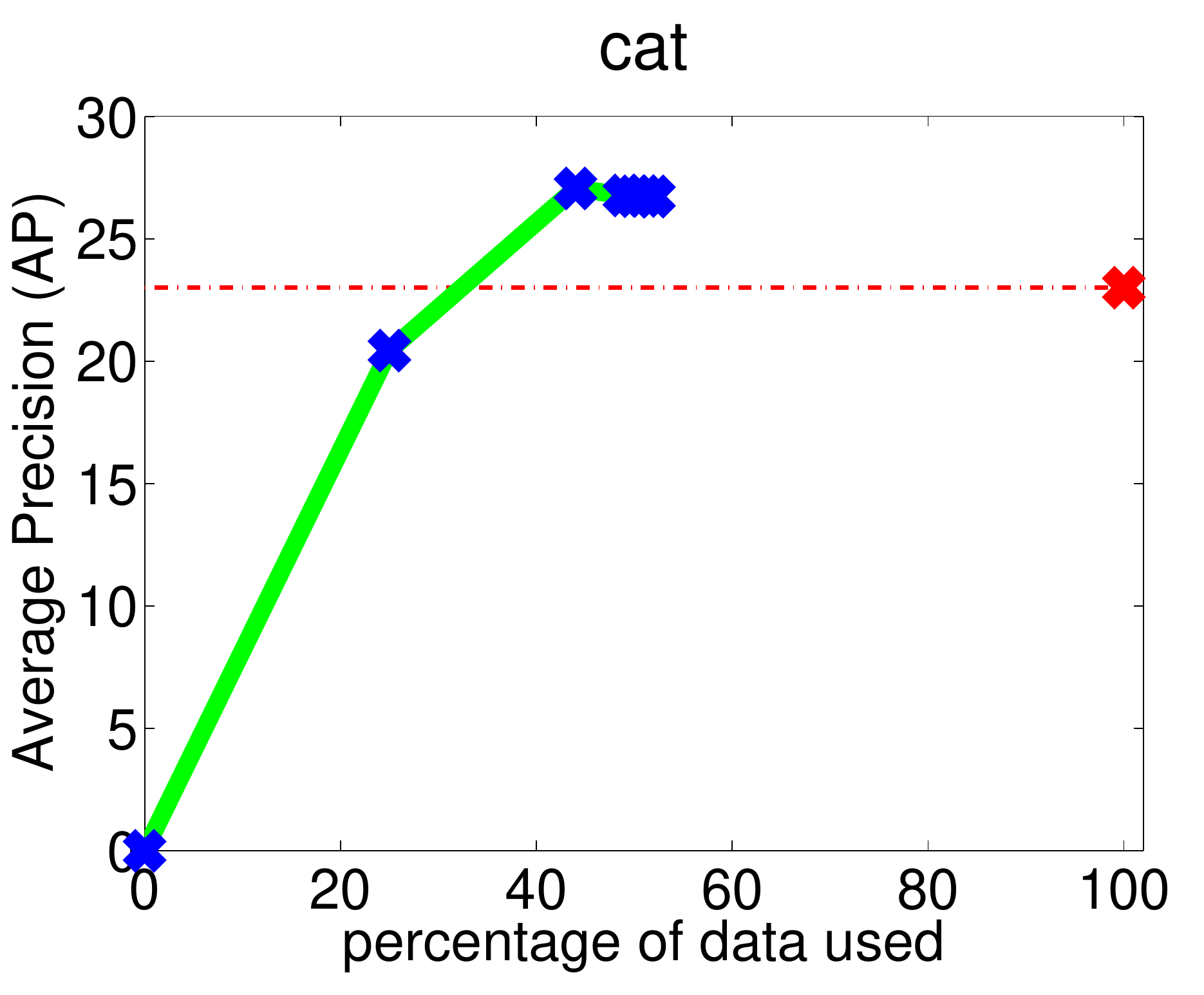}\\
  \includegraphics[clip=true, trim=1cm 6cm 1cm 8cm, width=0.21\textwidth]{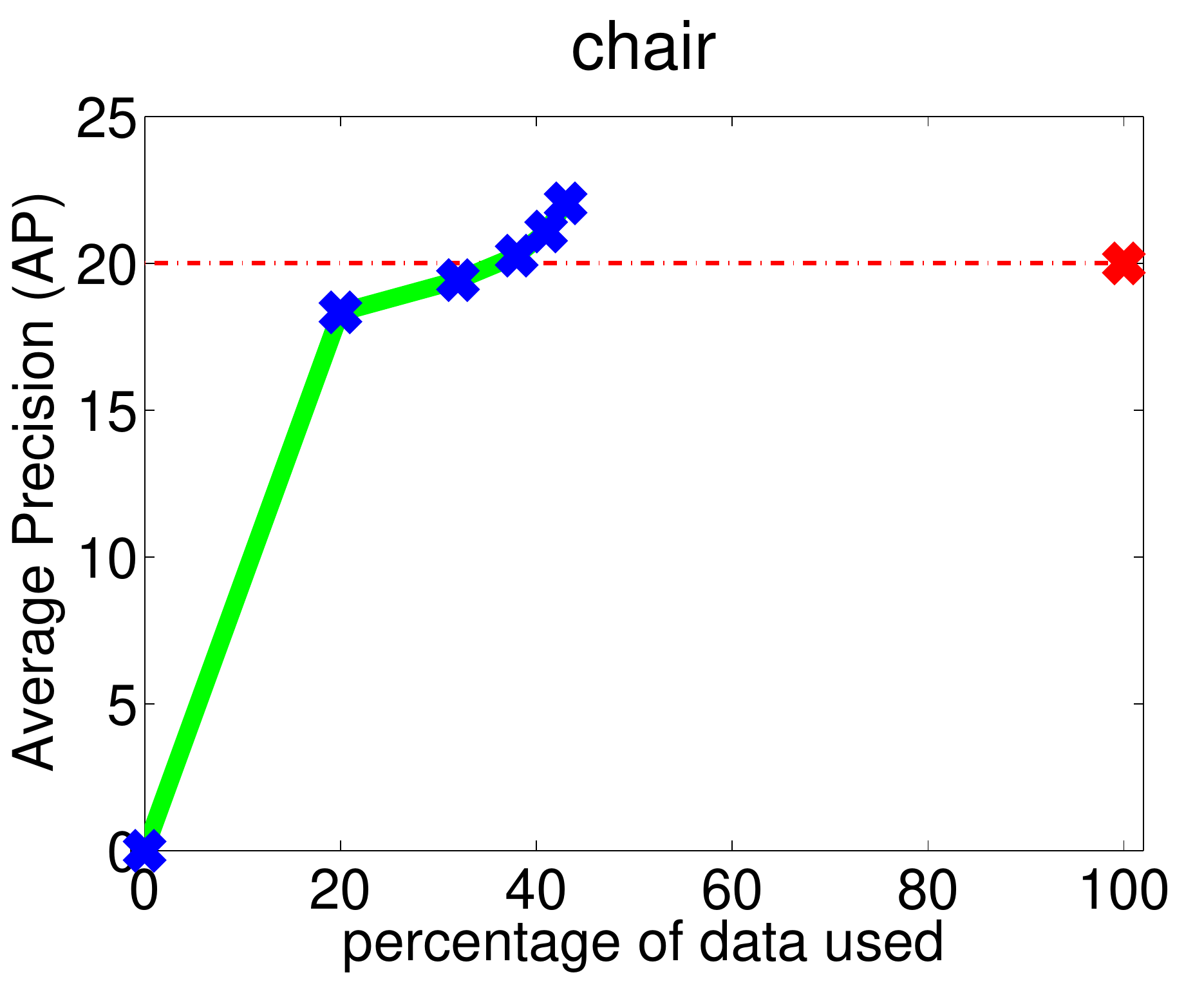}
  \includegraphics[clip=true, trim=1cm 6cm 1cm 8cm, width=0.21\textwidth]{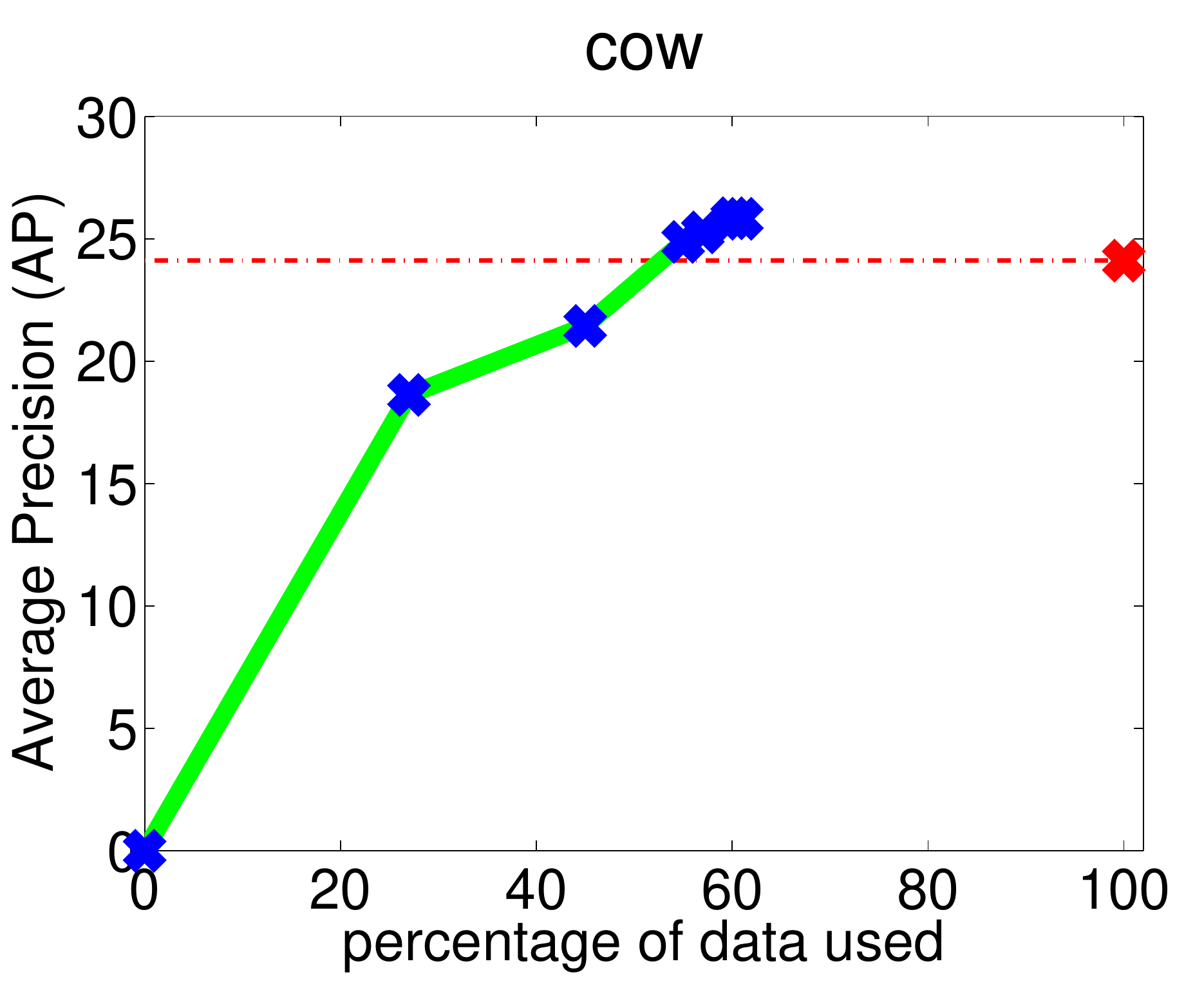}
  \includegraphics[clip=true, trim=1cm 6cm 1cm 8cm, width=0.21\textwidth]{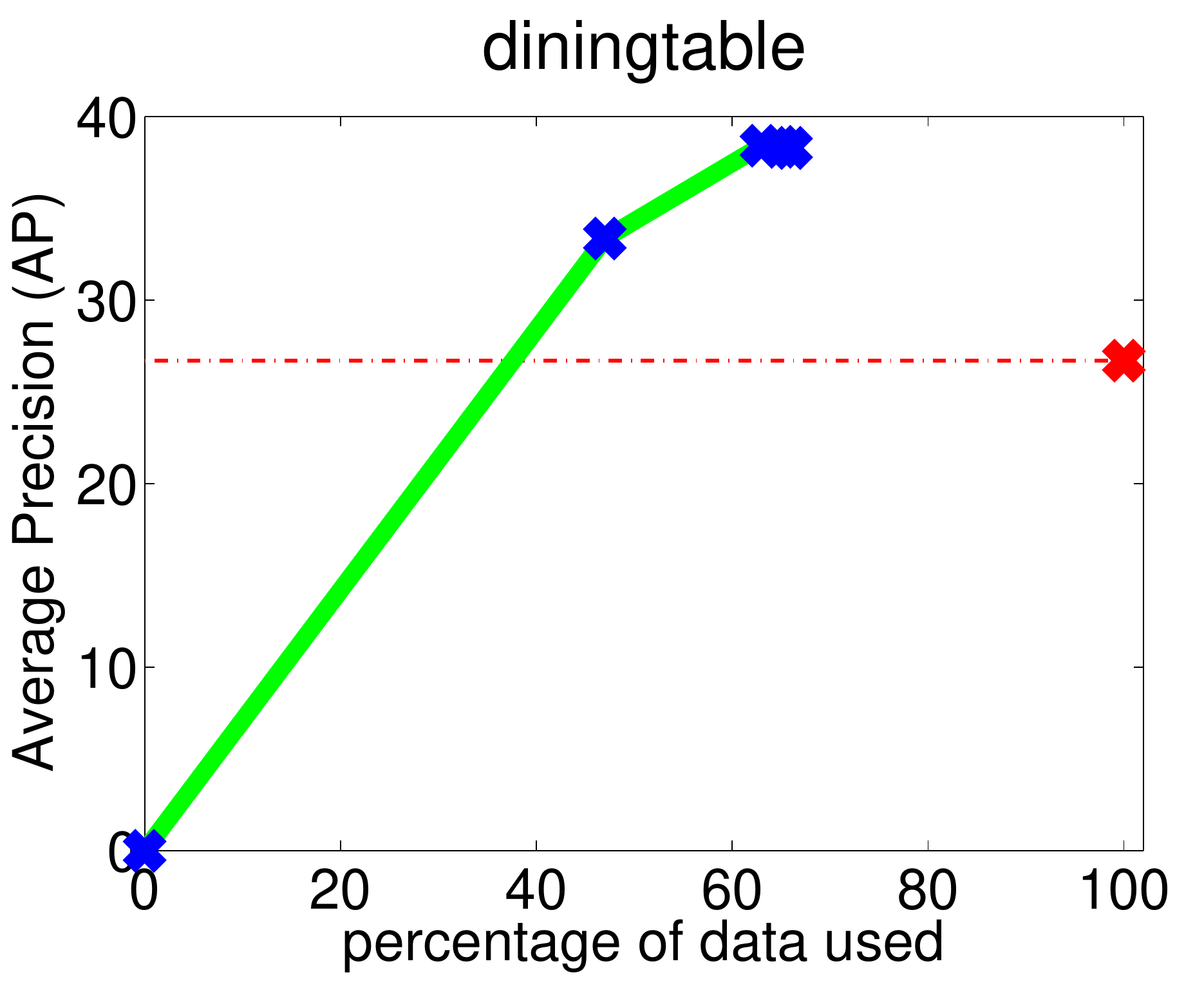}
  \includegraphics[clip=true, trim=1cm 6cm 1cm 8cm, width=0.21\textwidth]{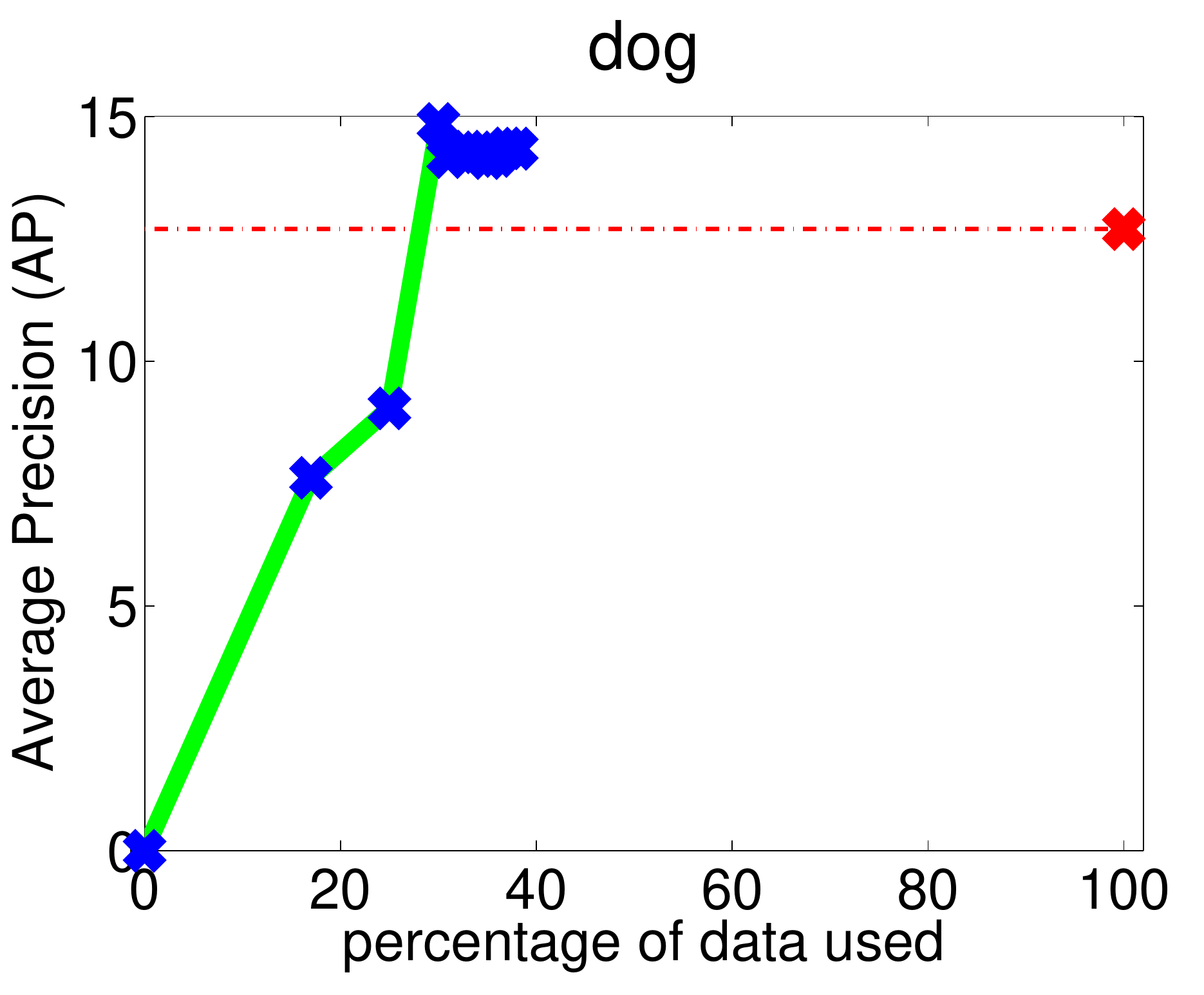}\\
  \includegraphics[clip=true, trim=1cm 6cm 1cm 8cm, width=0.21\textwidth]{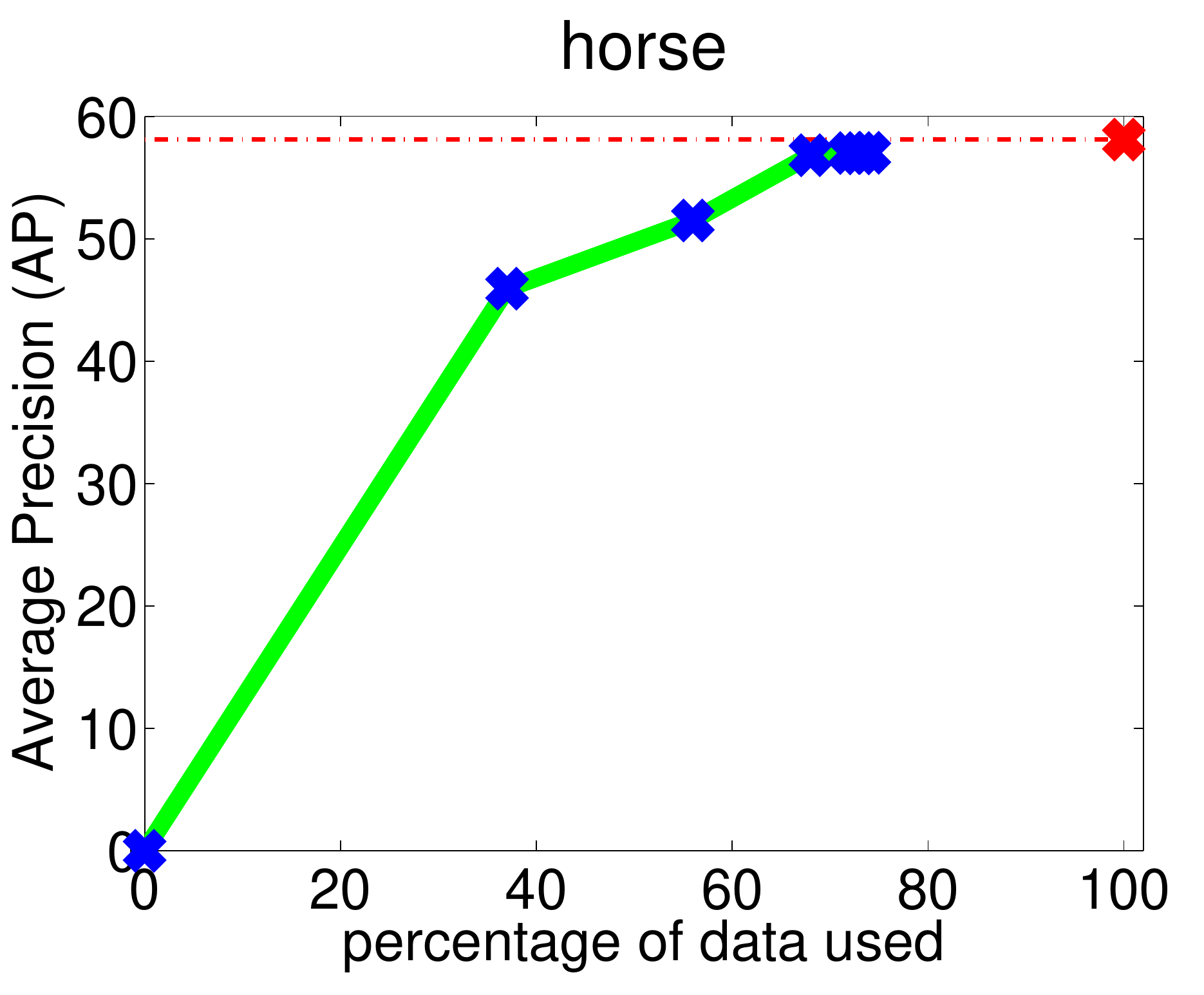}
  \includegraphics[clip=true, trim=1cm 6cm 1cm 8cm, width=0.21\textwidth]{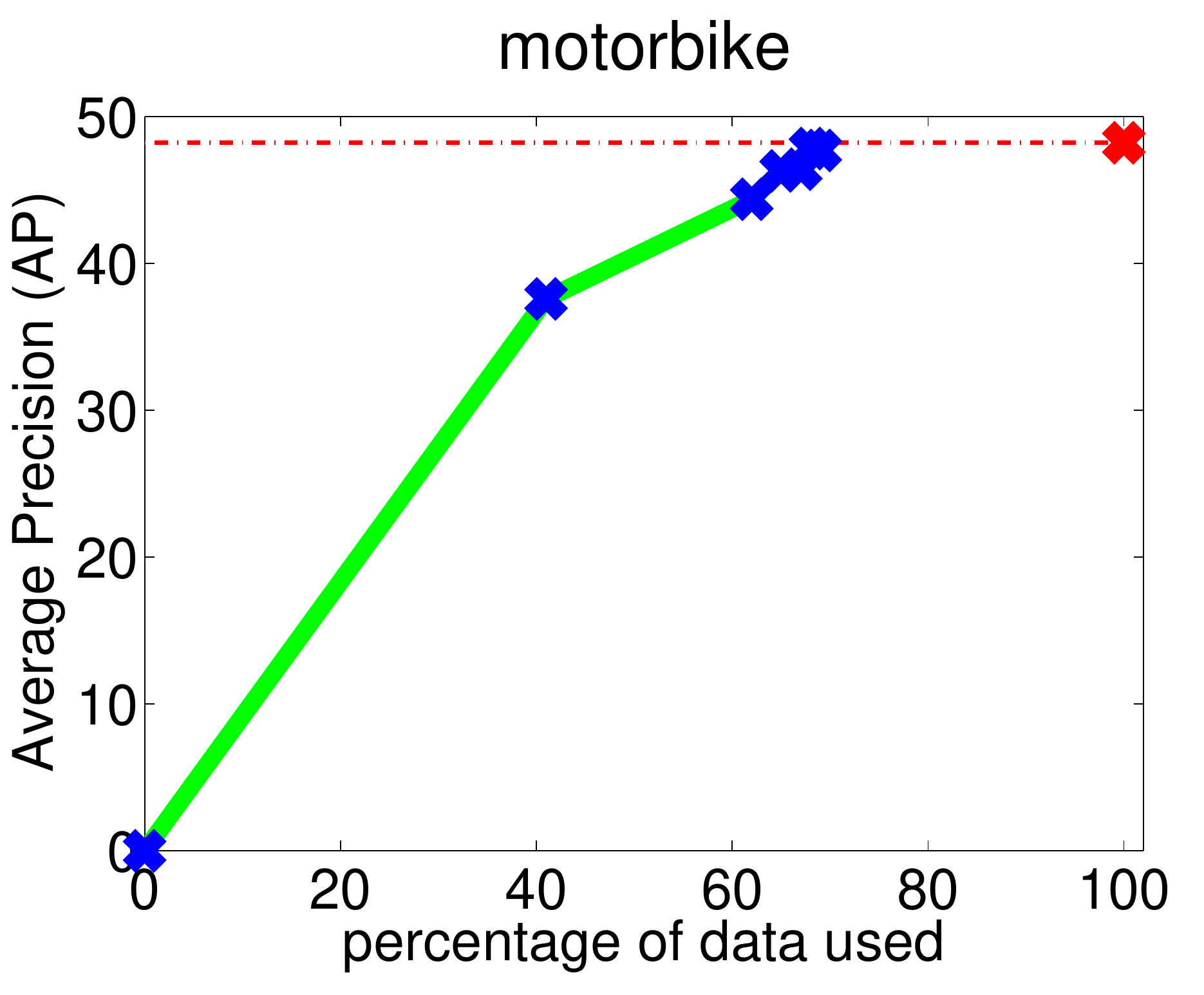}
  \includegraphics[clip=true, trim=1cm 6cm 1cm 8cm, width=0.21\textwidth]{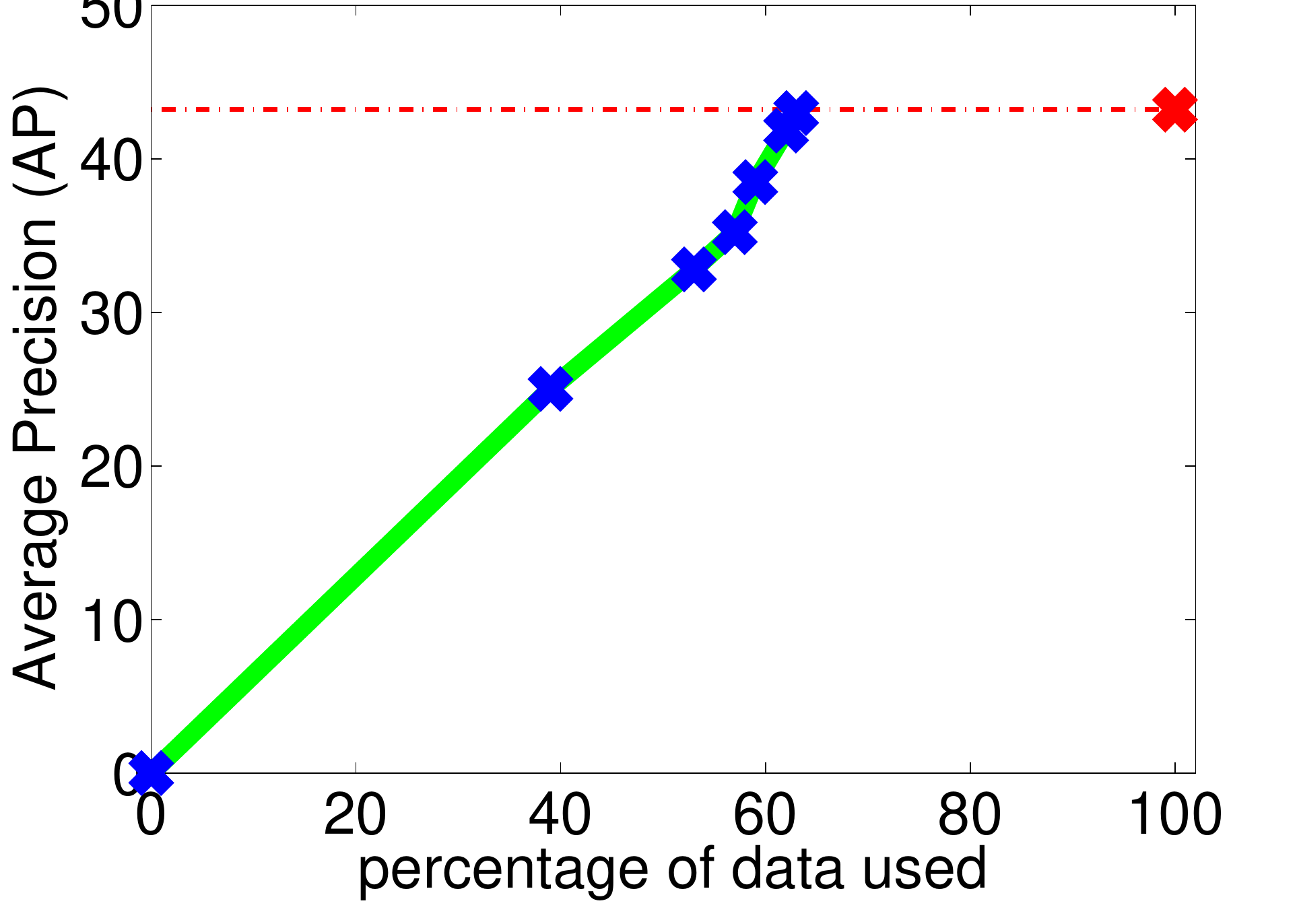}
  \includegraphics[clip=true, trim=1cm 6cm 1cm 8cm, width=0.21\textwidth]{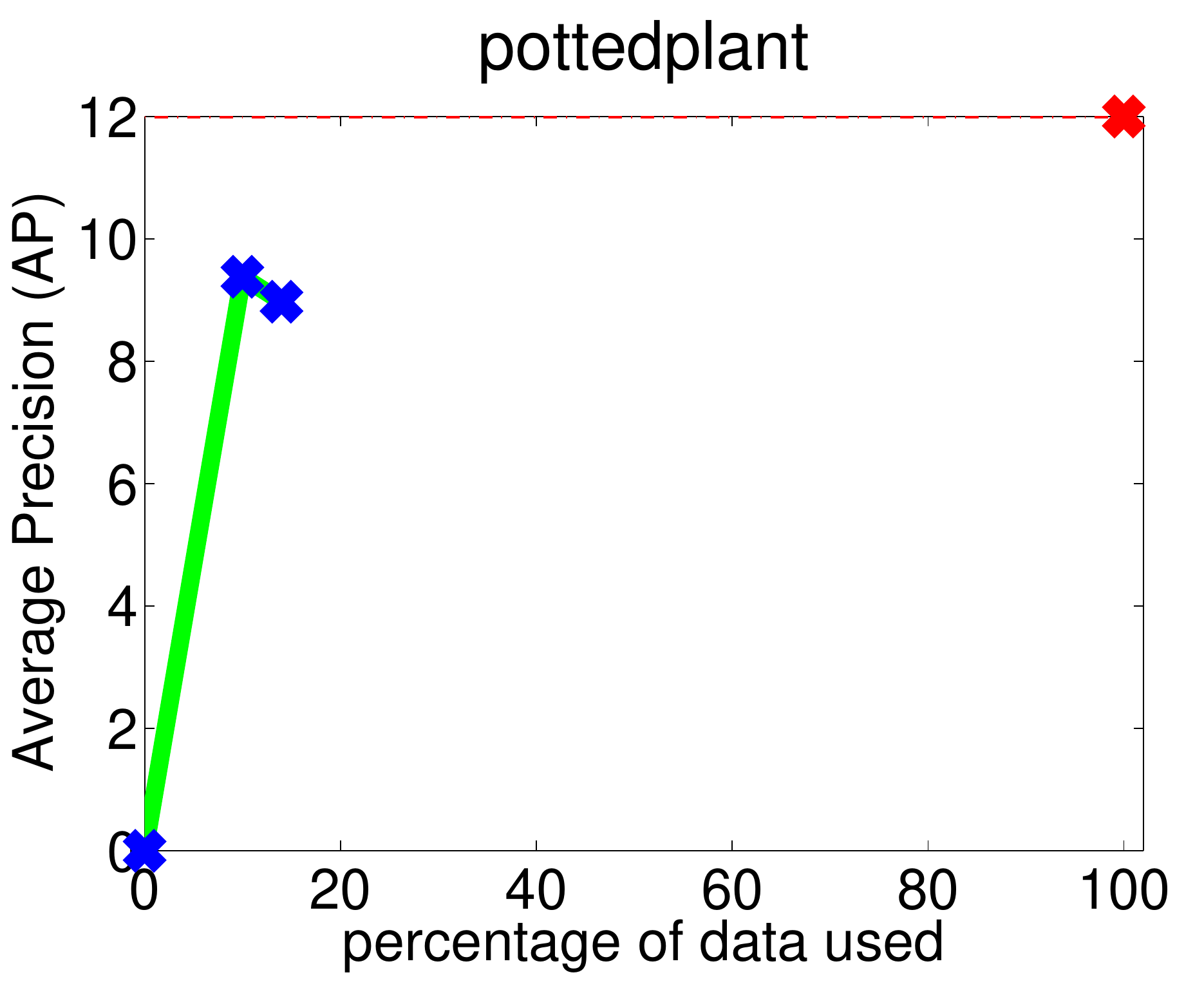}\\
  \includegraphics[clip=true, trim=1cm 6cm 1cm 8cm, width=0.21\textwidth]{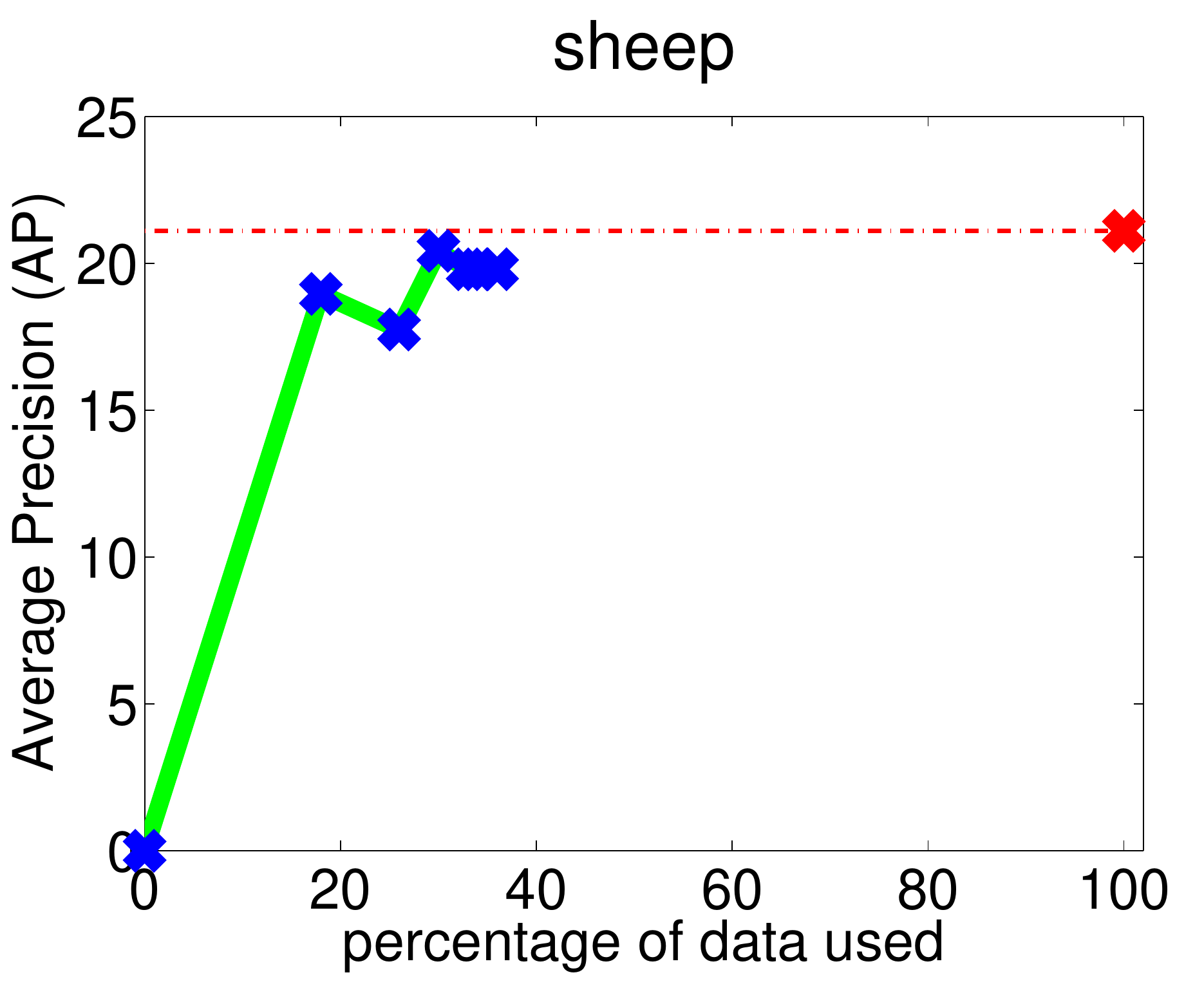}
  \includegraphics[clip=true, trim=1cm 6cm 1cm 8cm, width=0.21\textwidth]{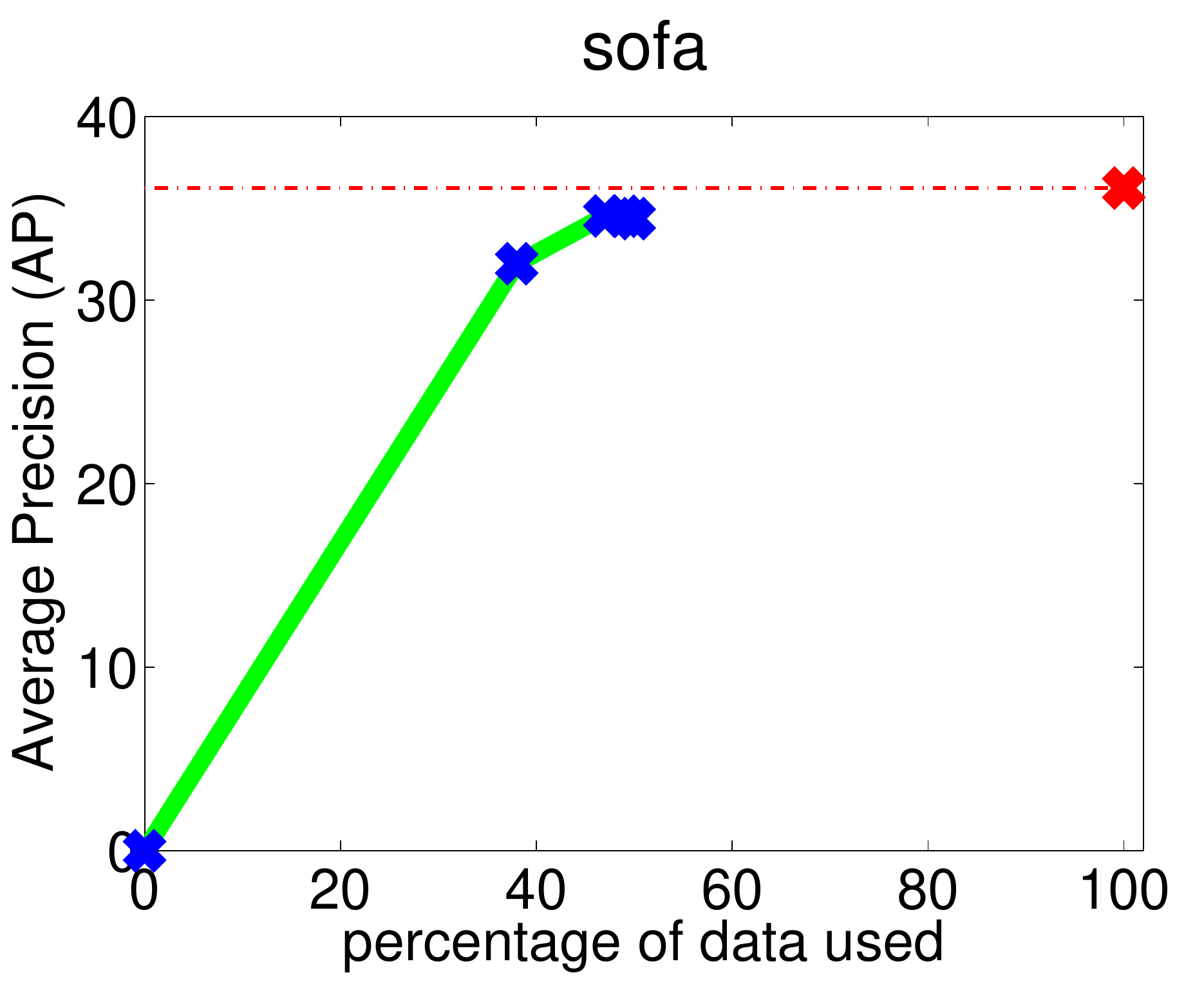}
  \includegraphics[clip=true, trim=1cm 6cm 1cm 8cm, width=0.21\textwidth]{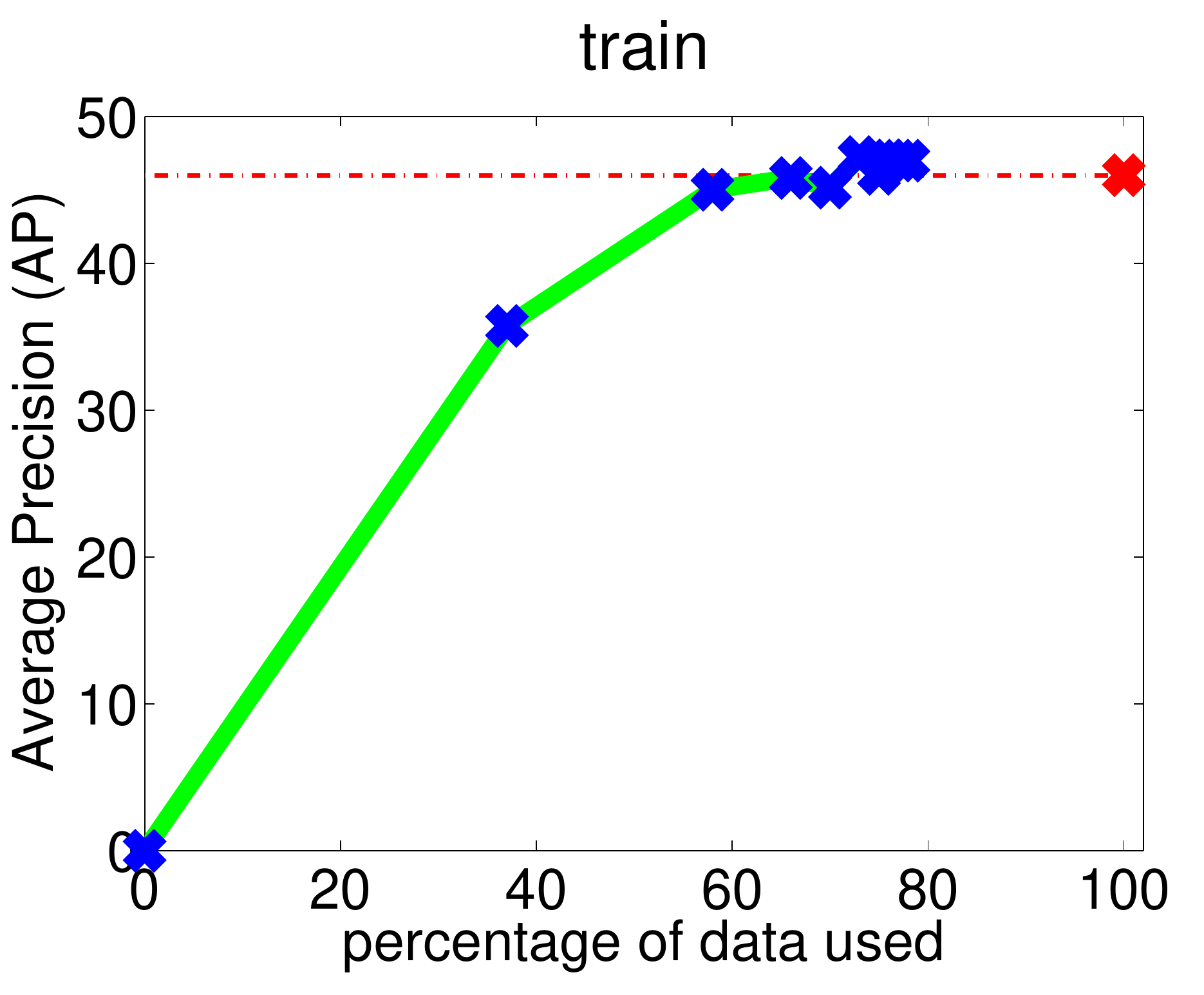}
  \includegraphics[clip=true, trim=1cm 6cm 1cm 8cm, width=0.21\textwidth]{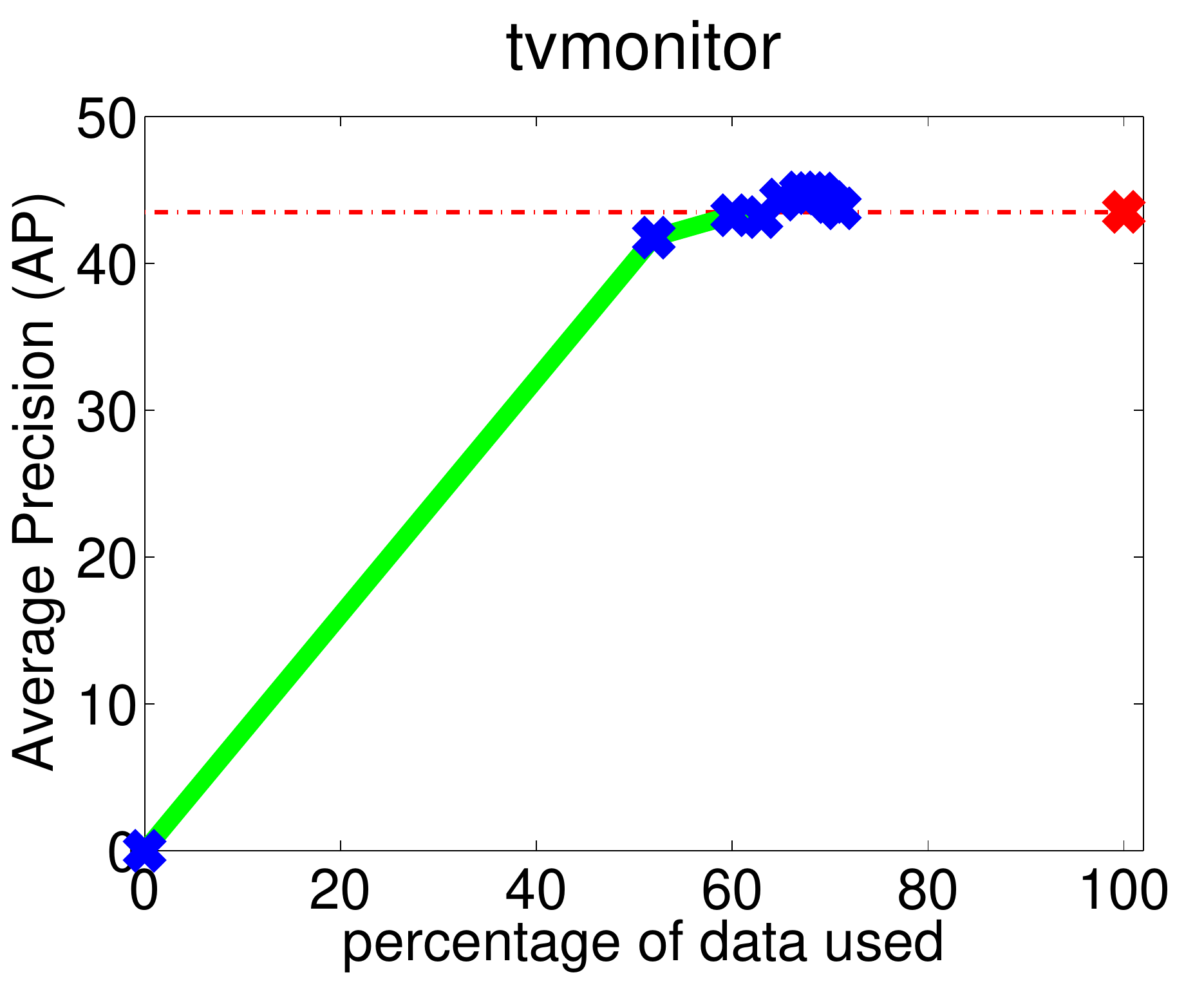}\\

  \caption{Evolution of the AP with more experts as a function of the fraction of the data used. Blue crosses in green (light gray) curve mark our method's AP after aggregation of each expert. Red cross (dashed line) is the AP of DPM using all data.}
  \label{fig:APrev}
\end{figure*}

To better understand the effect of a proper aggregation, we follow the work of Hoiem et al. in \cite{Hoiem12} and remove the false positives generated by poor localization. We remove the detections that overlap between 0.1 to 0.5 to the provided ground truth and re-compute the AP. We do the same procedure for our DPM baseline \cite{voc-release5}. Table (\ref{table:voc07noloc}) shows that we have a larger margin in the mean AP and we outperform DPM in 16 out of 20 classes using the same fraction of data as mentioned in Table (\ref{table:voc07}) for the "Ours" results. This indicates the significantly higher ability of our proposed classifiers in recalling the objects and indicates the benefit of designing an appropriate localization scheme for aggregation of the different experts to reduce the false positives due to multiple counting.  
Furthermore, Fig. (\ref{fig:APrev}) shows that in most of the classes after addition of the 5th mixture the performance does not necessarily increase. Most of the performance is gained by addition of the first experts. This highlights the importance of score calibration as more mixtures are added \cite{Malisiewicz11}.
\begin{table*}[!t]
\scriptsize
\centering
\tabcolsep=0.15cm
\begin{tabular}{ l c c c c c c c c c c c c c c c c c c c c c }
     \Xhline{5\arrayrulewidth}
     &aero&bike&bird&boat&bottle&bus&car&cat&chair&cow&table&
     dog&horse&mbike&person&plant&sheep&sofa&train&tv&mAP \\ 
     \rowcolor[gray]{0.8}
    DPM v5\cite{voc-release5}&37.3&62.2&10.4&\textbf{17.8}&\textbf{28.1}&56.8&\textbf{60.5}&27.9&22.0&24.8&30.0&13.4
     &\textbf{61.0}&49.5&52.4&14.3&21.9&38.2&51.7&44.3&36.2\\
     \rowcolor[gray]{0.8}
    Ours&\textbf{39.5}&\textbf{67.0}&\textbf{13.4}&16.4&22.7&\textbf{60.4}&58.5&\textbf{38.5}&\textbf{24.9}&\textbf{27.1}&\textbf{42.3}&\textbf{15.3}&59.6
    &\textbf{49.6}&\textbf{52.7}&\textbf{15.0}&\textbf{22.5}&\textbf{39.3}&\textbf{52.0}&\textbf{46.8}&\textbf{38.2}\\ 
 	 \hline    
\end{tabular}
\caption{Average Precision results on VOC07 by taking out the localization false positives. As suggested by \cite{Hoiem12} we remove detections that have between 0.1 to 0.5 overlap with the corresponding ground truth. The results are by using the same fraction of the data as mentioned in Table (\ref{table:voc07})}
\label{table:voc07noloc}
\end{table*}
\subsection{Sharing Samples}
On a subset of classes we show the evolution of the AP when adding more mixtures by comparing with the case when we do not allow sharing (Fig. \ref{fig:APrevNog}). The gain in the improvement when sharing samples between mixtures indicate the importance of this procedure, possibly because it generates smoother boundaries and help the generalization of each of the experts for the final aggregation.
\subsection{Qualitative Results}
\begin{figure*}
  \mbox{}\\
  \centering
  \includegraphics[clip=true, trim=2cm 12cm 3cm 8cm, width=0.33\textwidth]{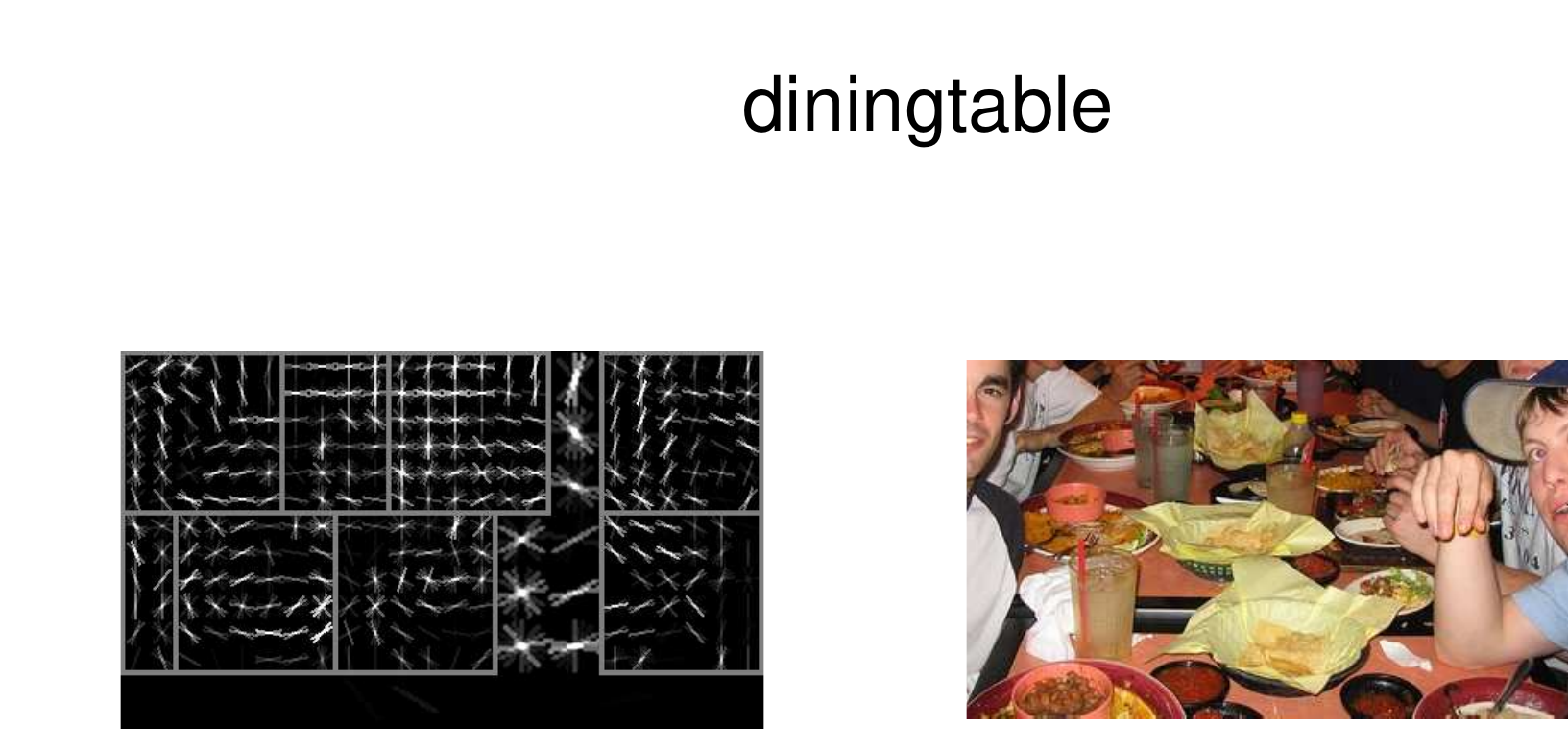}
  \includegraphics[clip=true, trim=2cm 11cm 3cm 8.5cm, width=0.33\textwidth]{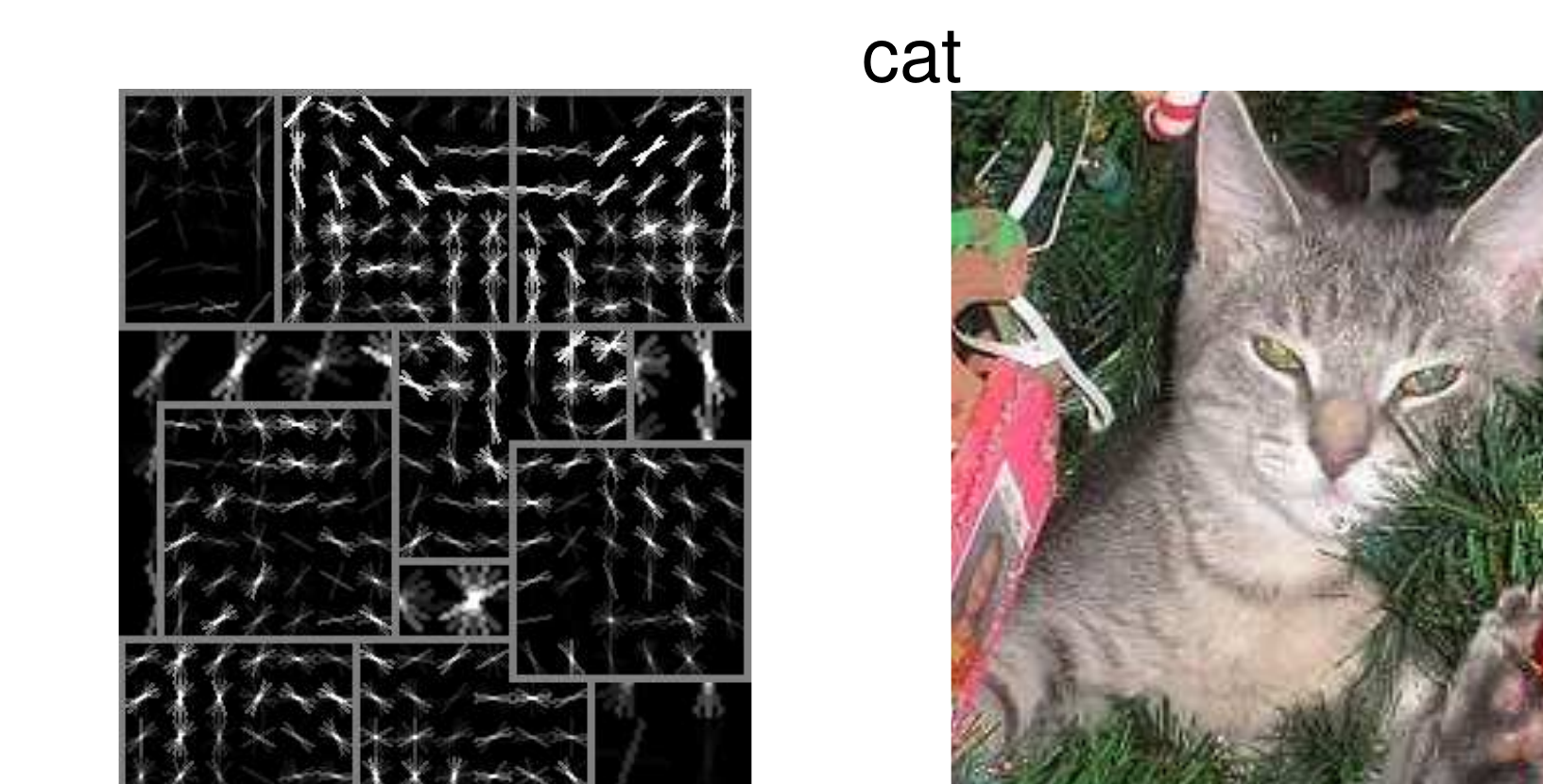}
  \includegraphics[clip=true, trim=2cm 11cm 3cm 8.5cm, width=0.33\textwidth]{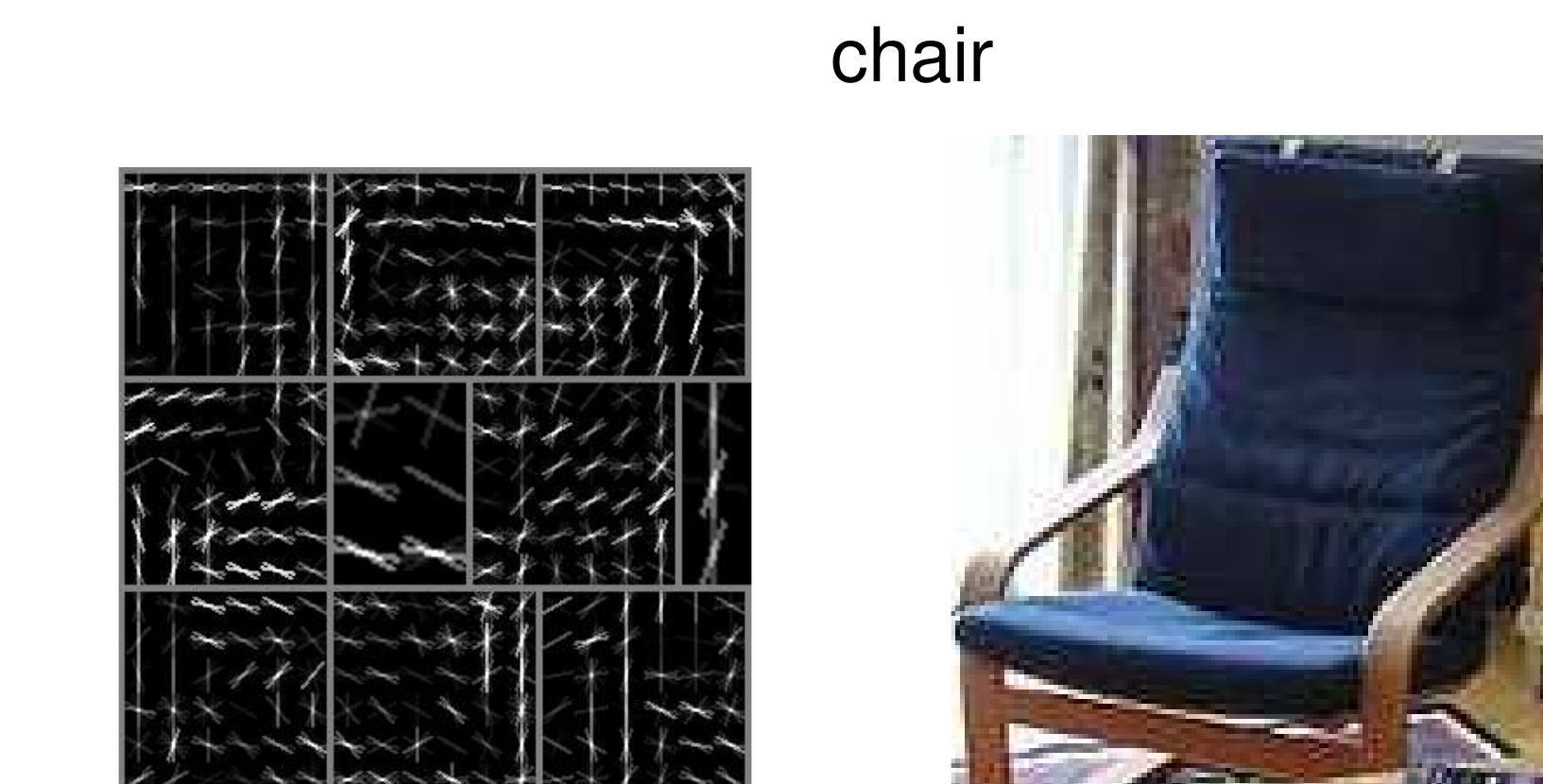}\\
    \includegraphics[clip=true, trim=2cm 8cm 3cm 8cm, width=0.18\textwidth]{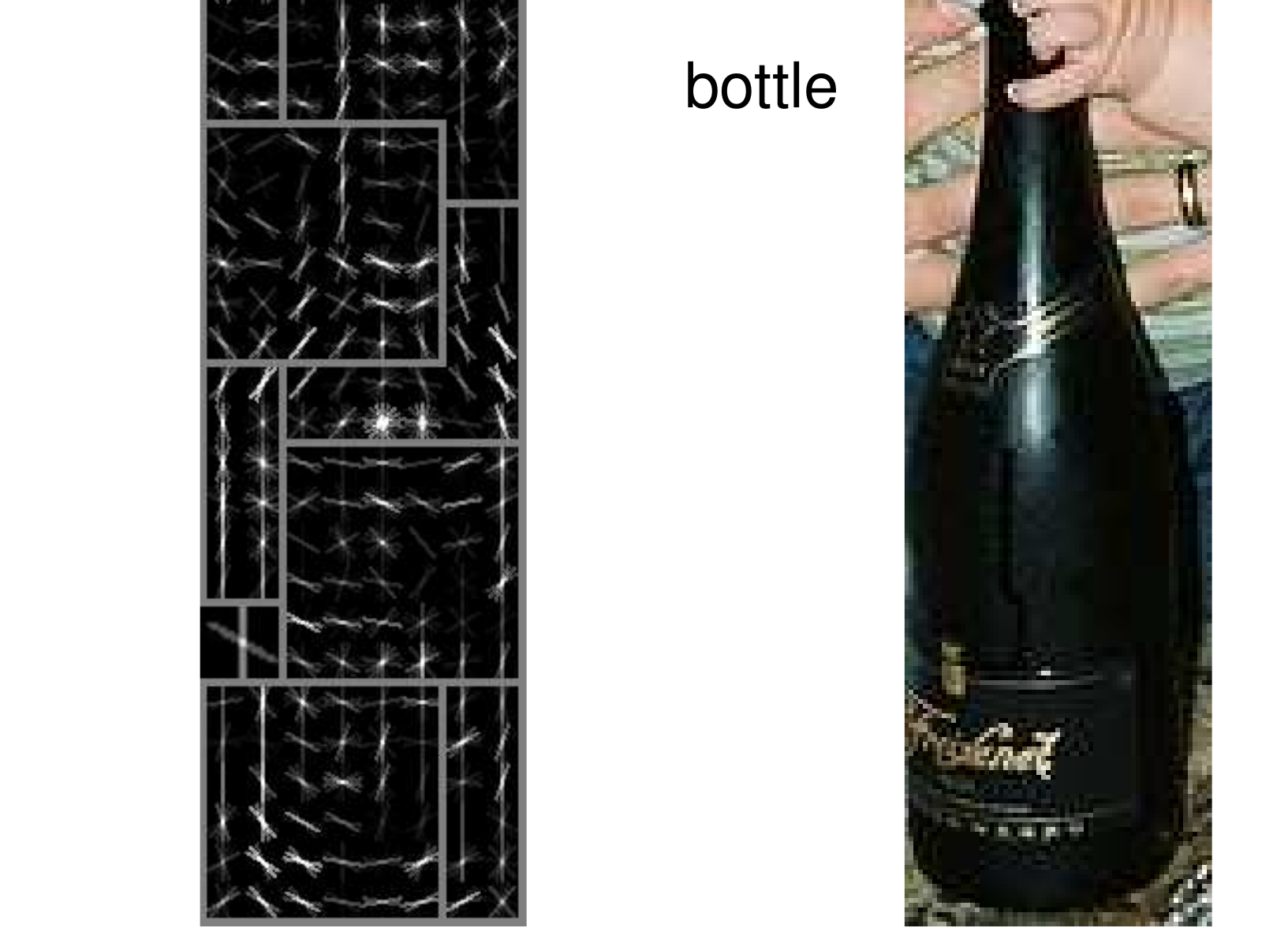}  
    \includegraphics[clip=true, trim=2cm 12cm 3cm 8cm, width=0.33\textwidth]{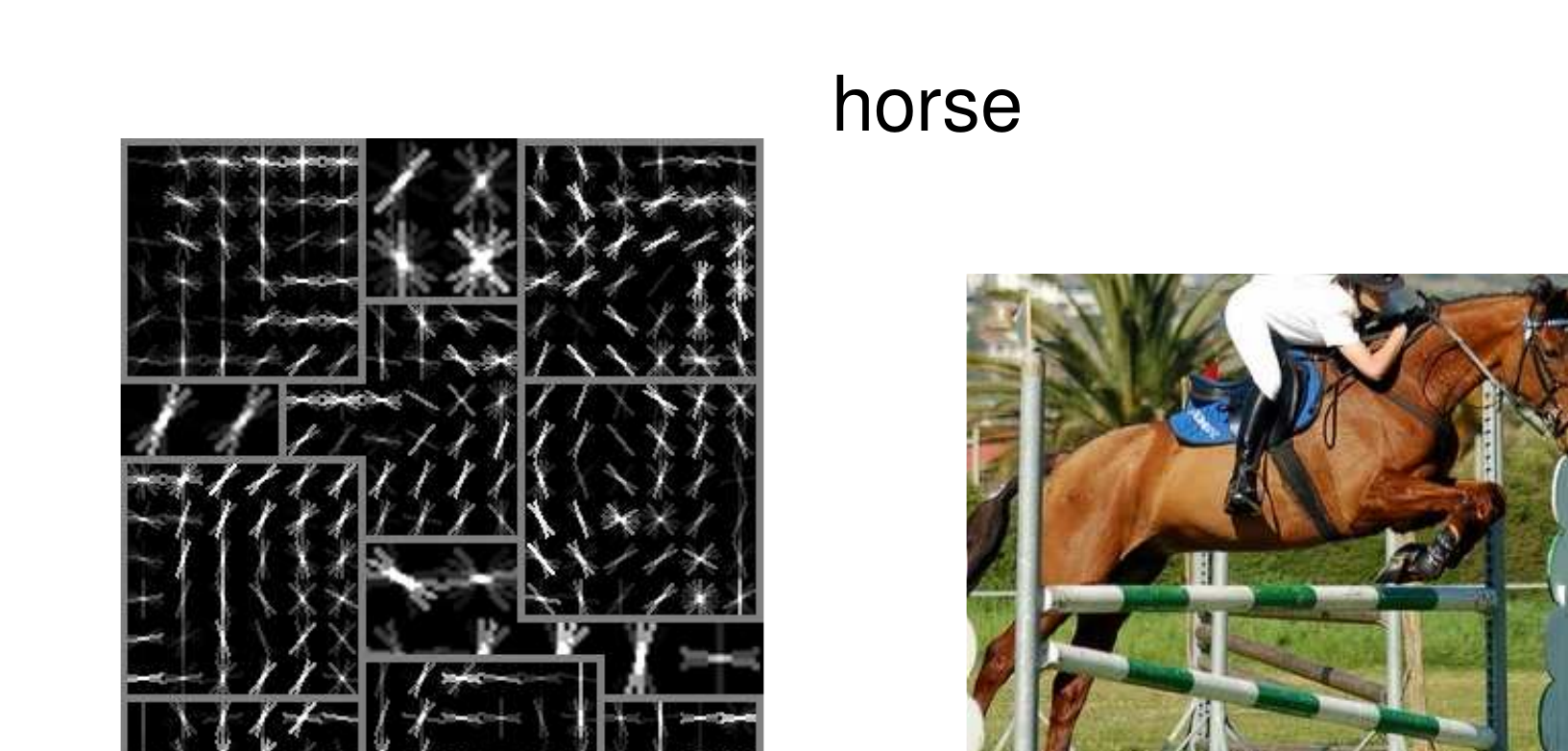}
        \includegraphics[clip=true, trim=2cm 12cm 3cm 9cm, width=0.33\textwidth]{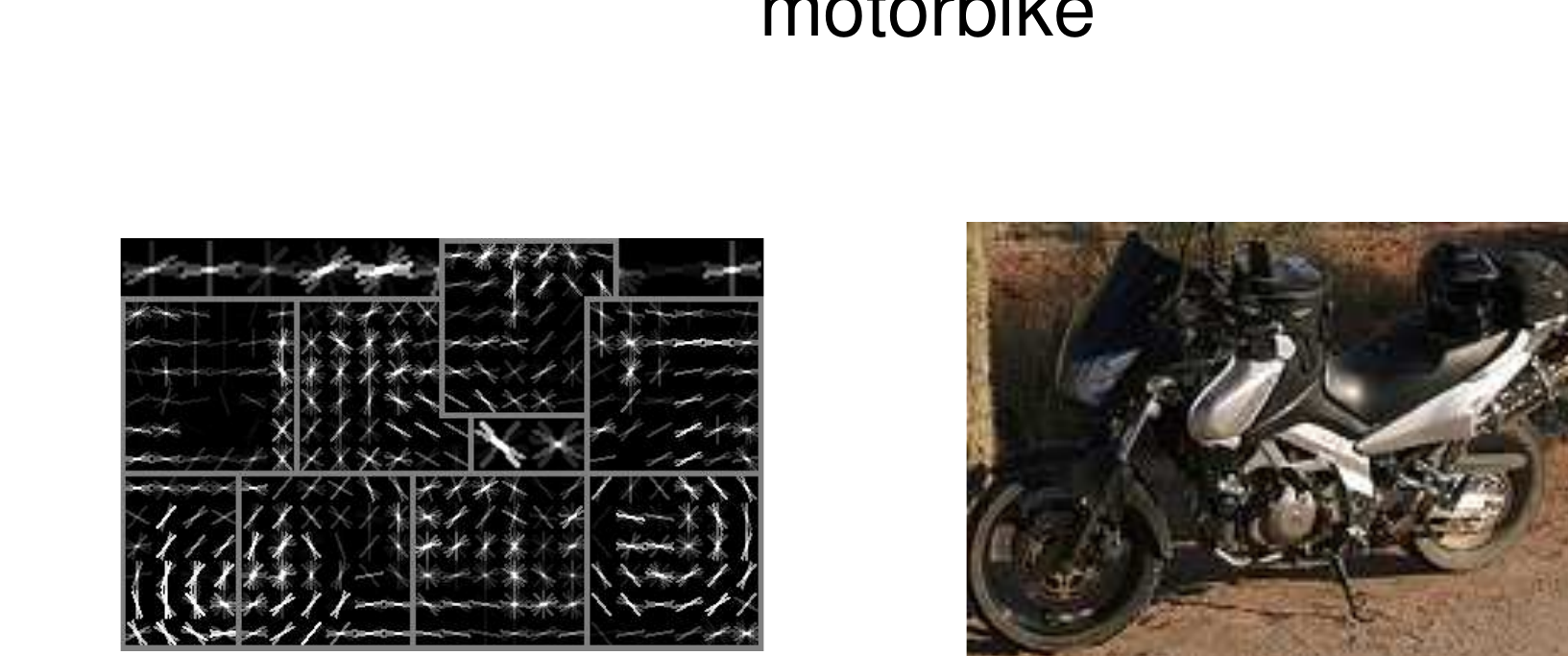}

  \caption{Trained filters for the largest visual subclass found for a subset of classes along with one of their members.}
  \label{fig:filters}
\end{figure*}

Fig. \ref{fig:filters} depicts the HOG filters and a sample for the core subset of each of the classes. The sharp and clear edges exhibited in the filters show the good alignment of the samples caused by removal of outliers in the core subclasses. \\ To see figures illustrating the samples assigned to each expert and how they share them among themselves (such as Figure \ref{fig:teaser}) please refer to supplementary material.
\subsection{Statistics of isolated exemplars and miss-detections}
\begin{figure*}
  \centering
\begin{tabular}{l|r}

  \includegraphics[clip=true, trim=0cm 0cm 0cm 0cm, width=0.35\textwidth]{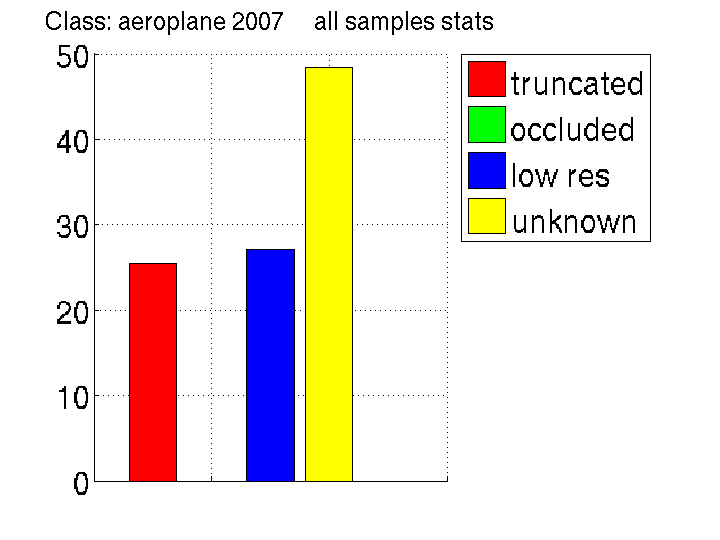}& 
  \includegraphics[clip=true, trim=0cm 0cm 0cm 0cm, width=0.35\textwidth]{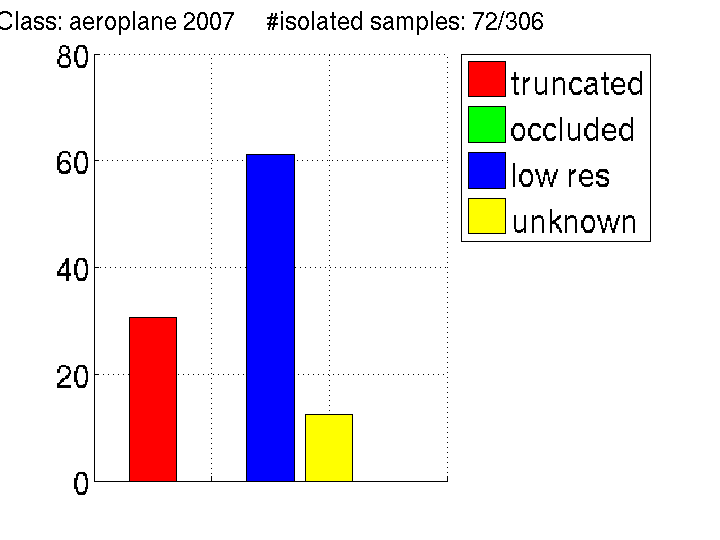}\\
    \includegraphics[clip=true, trim=0cm 0cm 0cm 0cm, width=0.35\textwidth]{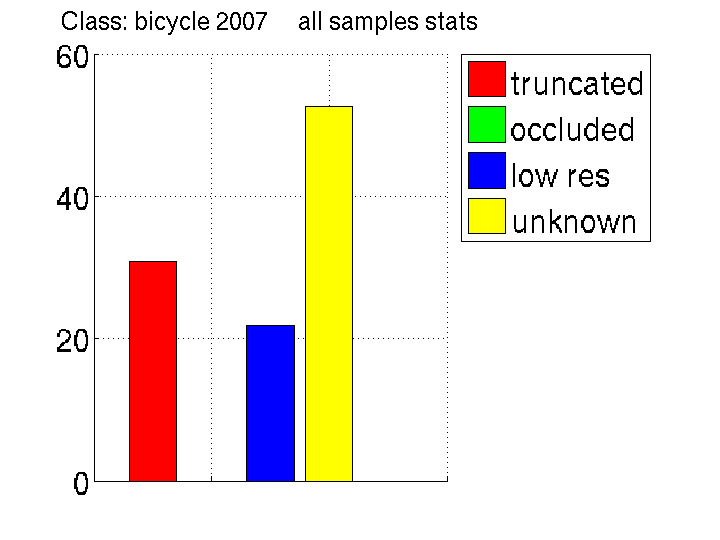}& 
    \includegraphics[clip=true, trim=0cm 0cm 0cm 0cm, width=0.35\textwidth]{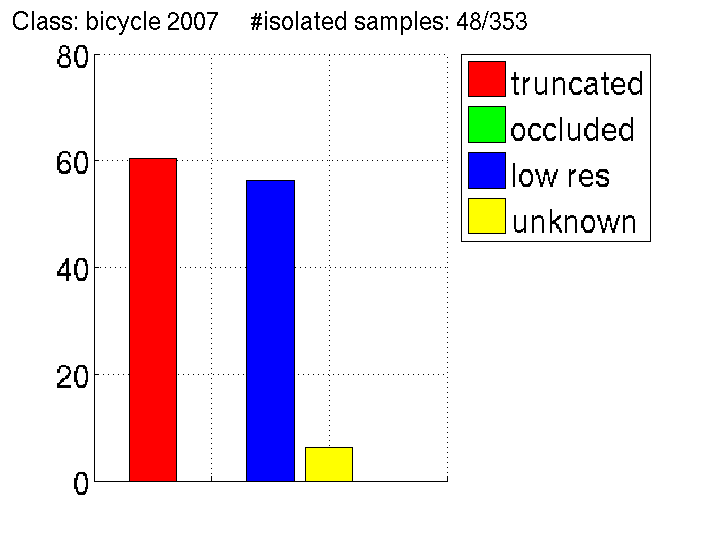}\\
    \includegraphics[clip=true, trim=0cm 0cm 0cm 0cm, width=0.35\textwidth]{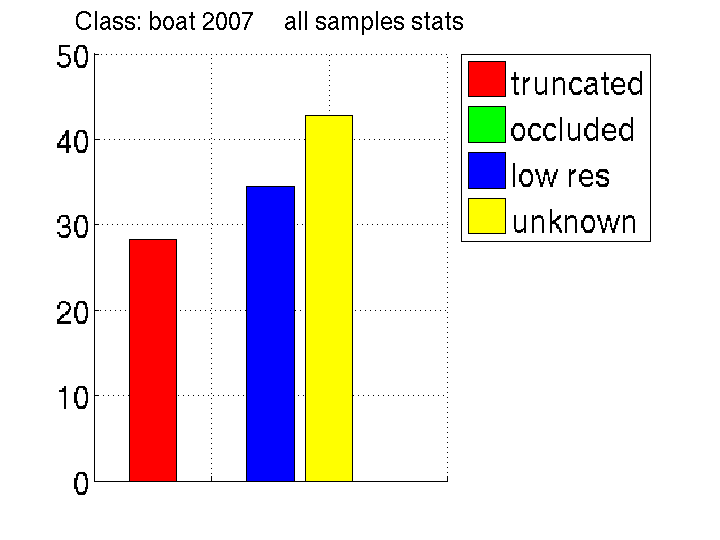}& 
    \includegraphics[clip=true, trim=0cm 0cm 0cm 0cm, width=0.35\textwidth]{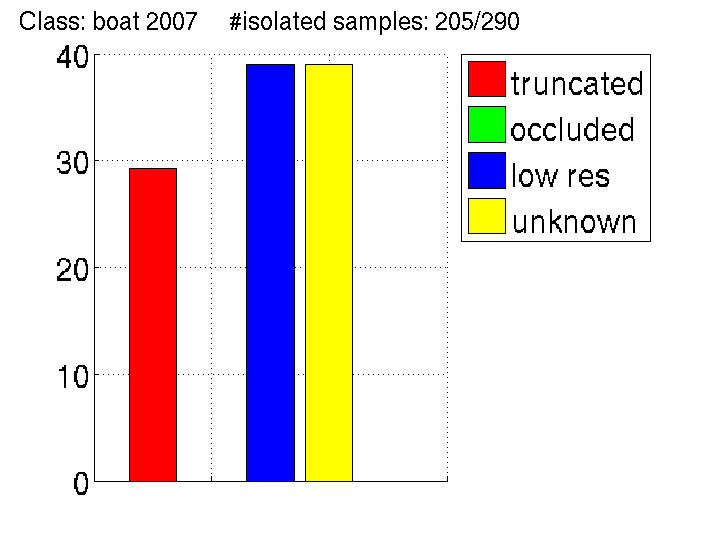}\\
    \includegraphics[clip=true, trim=0cm 0cm 0cm 0cm, width=0.35\textwidth]{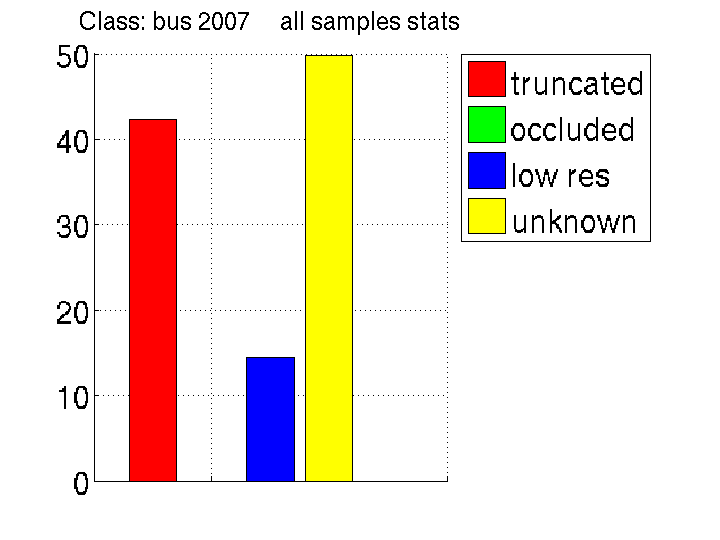}&
    \includegraphics[clip=true, trim=0cm 0cm 0cm 0cm, width=0.35\textwidth]{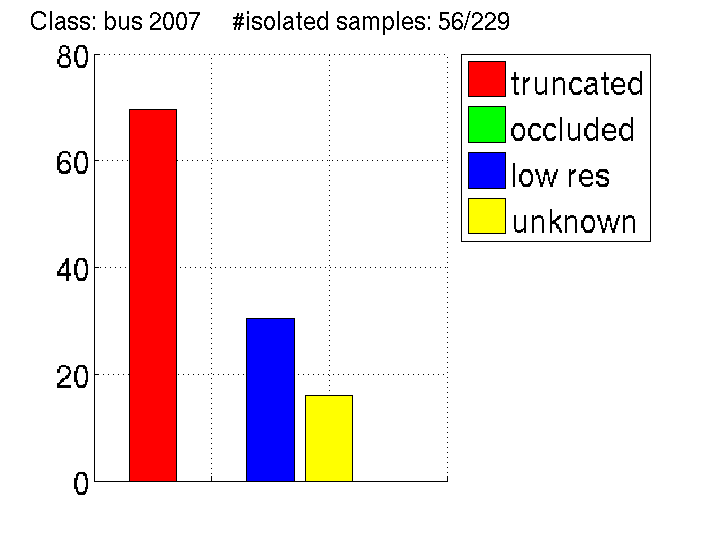}\\
    \includegraphics[clip=true, trim=0cm 0cm 0cm 0cm, width=0.35\textwidth]{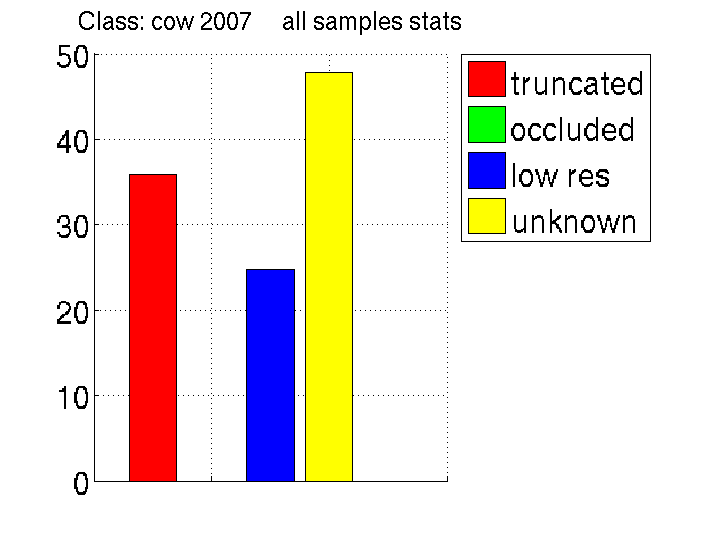}&
    \includegraphics[clip=true, trim=0cm 0cm 0cm 0cm, width=0.35\textwidth]{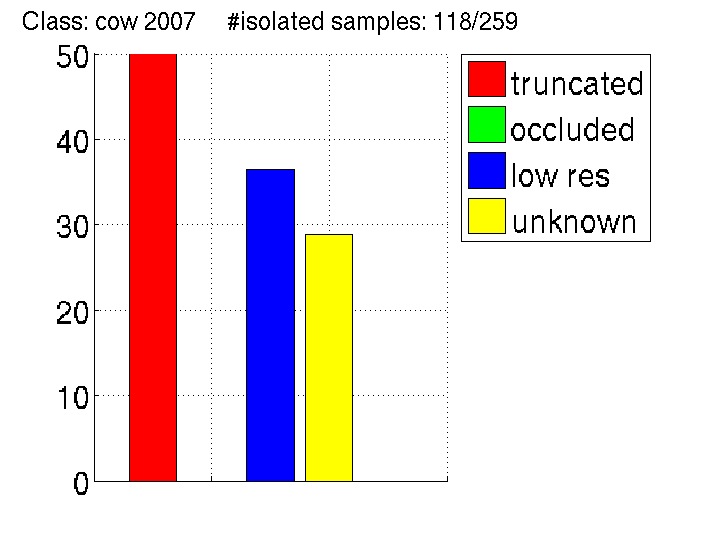}\\
  \end{tabular}
  \caption{Statistics on the distribution of isolated samples (members of clusters with size less than 5). The distribution is shown on three criteria: low resolution, truncated, unknown}
  \label{fig:isolated}
\end{figure*}

\begin{figure*}
  \centering
  \includegraphics[clip=true, trim=0cm 0cm 0cm 0cm, width=0.4\textwidth]{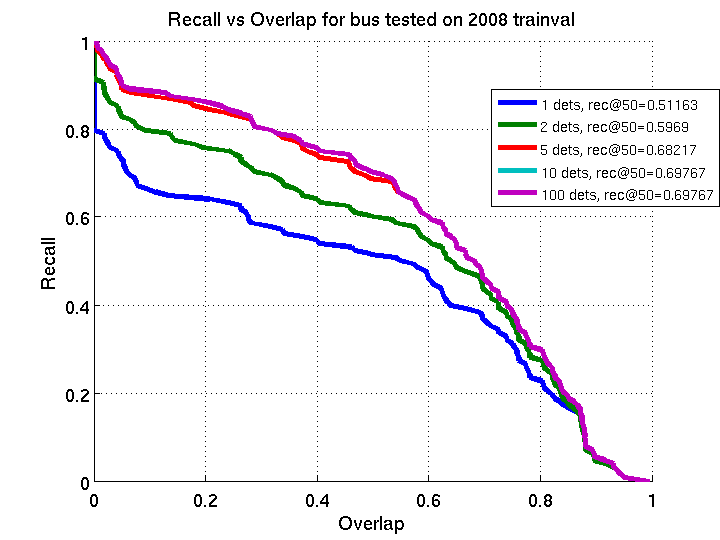} 
  \includegraphics[clip=true, trim=0cm 0cm 0cm 0cm, width=0.4\textwidth]{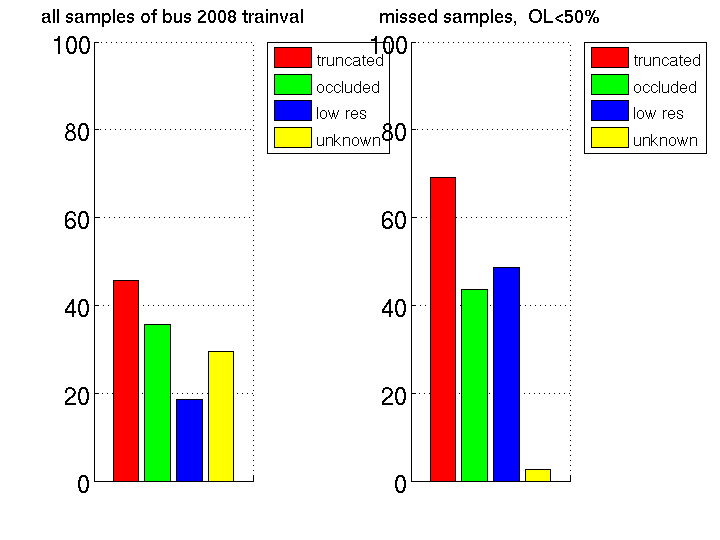} \\
  \includegraphics[clip=true, trim=0cm 0cm 0cm 0cm, width=0.4\textwidth]{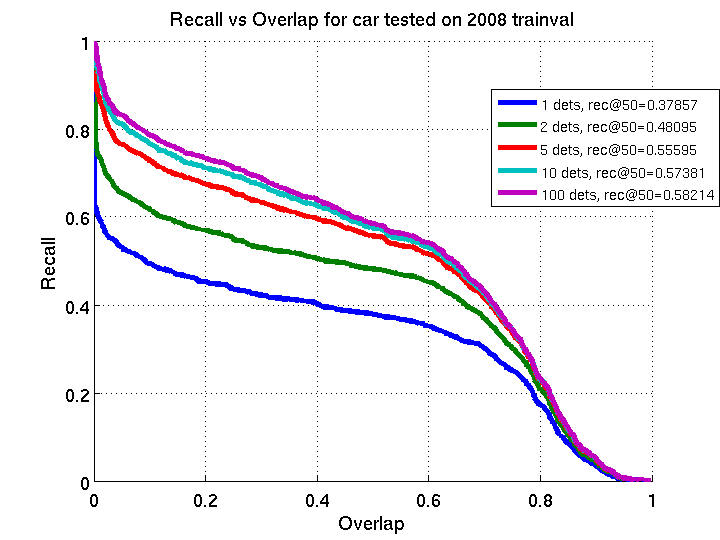} 
  \includegraphics[clip=true, trim=0cm 0cm 0cm 0cm, width=0.4\textwidth]{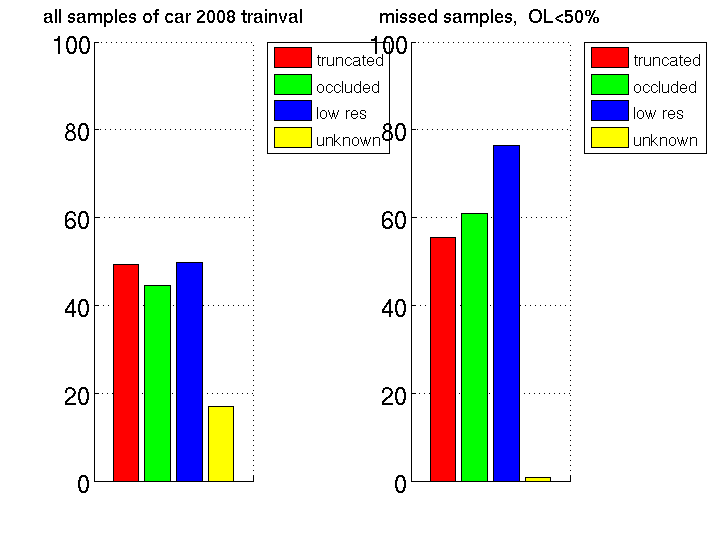} \\
  \includegraphics[clip=true, trim=0cm 0cm 0cm 0cm, width=0.4\textwidth]{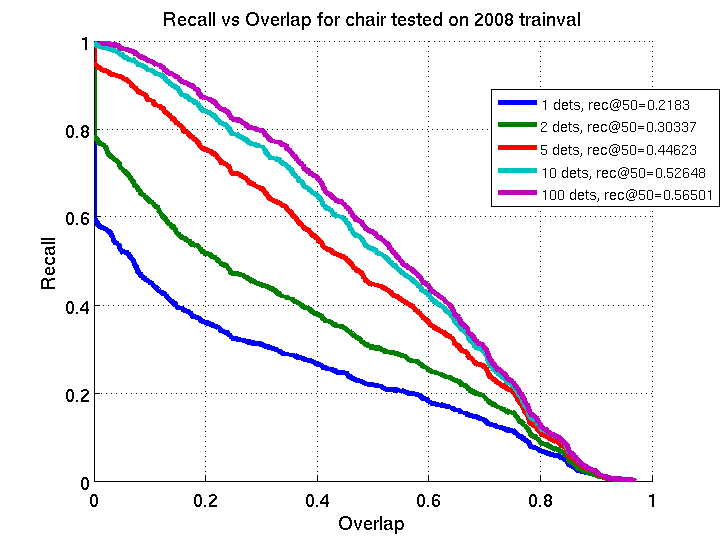} 
  \includegraphics[clip=true, trim=0cm 0cm 0cm 0cm, width=0.4\textwidth]{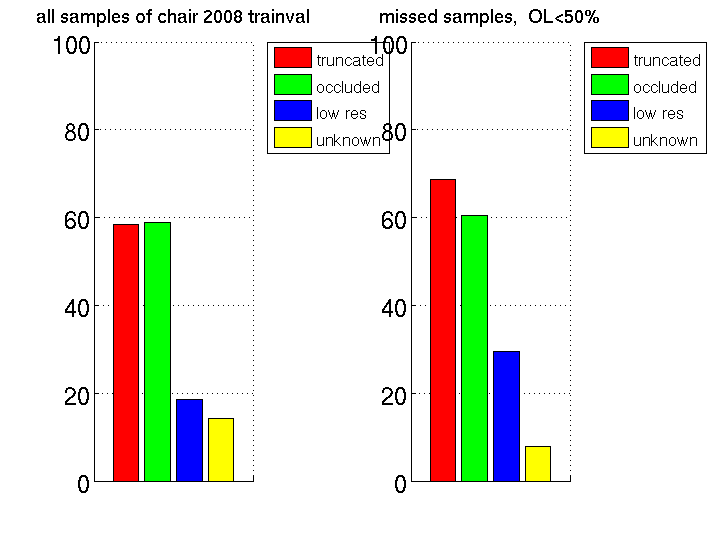} \\
  \includegraphics[clip=true, trim=0cm 0cm 0cm 0cm, width=0.4\textwidth]{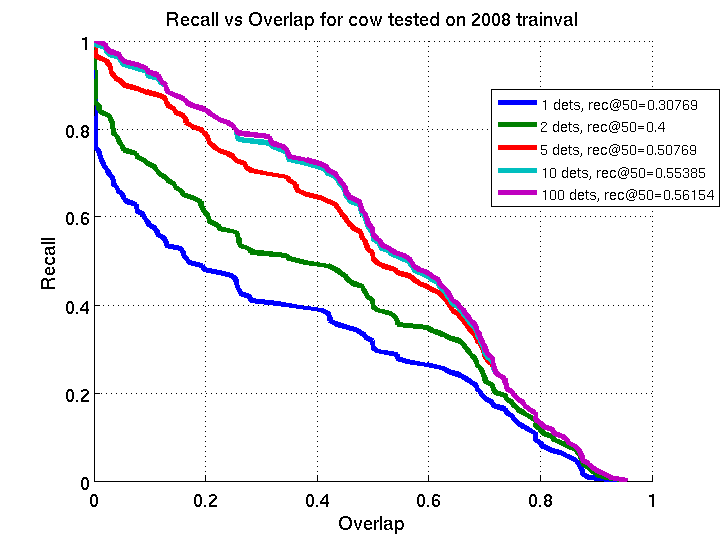} 
  \includegraphics[clip=true, trim=0cm 0cm 0cm 0cm, width=0.4\textwidth]{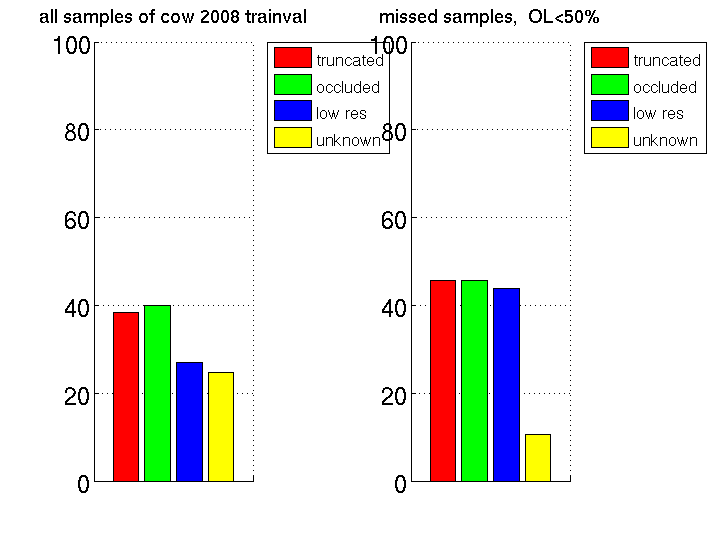}\\
  \caption{Statistics on the distribution of missed detection on PASCAL VOC 2008 training set. The distribution is shown on four criteria: low resolution, truncated, occluded, unknown. It can be seen that we have a relatively very low number of missed samples which are neither low resolution, truncated nor occluded.}
  \label{fig:2008}
\end{figure*}

We take a closer look into the discovered clusters. First, we try to understand the reason behind samples which end up in isolated exemplars. For this purpose we study the clusters with size less than 5. Then using the PASCAL annotations we compare their distribution (in terms of size and truncation) to the distribution of all training samples. It can be seen in figure \ref{fig:isolated} that the distribution of the isolated exemplars is biased toward less "unknown" samples. "unknown" means the sample is neither low resolution nor truncated.\\
Furthermore we study the statistics of the missed recalls on the PASCAL VOC \textbf{2008} dataset. Figure \ref{fig:2008} depicts the recall-overlap plots for (1,2,5,10,100) highest scoring windows of each training image. Then we pinpoint the missed samples in the 100 highest scoring windows and show the statistics of those missed samples compared to all samples. Again it can be observed that the low resolution and truncated samples are the most problematic ones. It can be seen that we have a relatively very low number of missed samples which are neither low resolution, truncated nor occluded.\\
From the two figures it can be observed that the samples at tails of the distribution in our representation (HOG features) are mostly comprised of low resolution, occluded and truncated samples.

\section{Conclusion and Future Works}
In this work we have proposed a novel method of designing classifiers that tries to incrementally discover and learn visual subclasses of the semantic class. The formulation is such that it tries to find largest subset of the training samples that can be reliably distinguished from the negative class. The initial attempt has shown that the state of the art object detectors are unable to use around half of the data (on average) appearing at the tails of the distribution, thus showing potentials for future works. In fact we showed the performance is slightly improved by taking the outliers (isolated exemplars) out of the training set.\\
Results by ignoring the poor localization false positives indicates that a suitable aggregation procedure for the output of the independently trained experts appears to be a promising future work to reduce false positives due to multiple counting. However, we think the most important future direction is the analysis of a way to enlarge or merge the discovered subclasses by including samples from the tails of the distribution. This can be done by re-designing the model, feature representations or by using additional data to fill in the identified gaps of the input space (between isolated exemplars).

\section{Appendix A.}
\label{sec:appendixA}
\subsection{Individual subclass discovery and learning}
We are interested in finding the \emph{largest subset} of our positive training data ($y=1$) which maintains a regular structure (large margin in case of L2 regularization) to \textit{all} the negative data ($y=-1$).
In the follwoing we first explain two alternative regularized risk minimization methods of largest subclass discovery. These methods use different approaches to  calculate the induced loss and are inspired by self-paced learning and regularized ramp loss learners respectively.
\small
\begin{align}
\label{eq_individual}
\underset{\mathbf{w},\mathbf{v}}{\text{minimize}}\,\,\dot{J}(\mathbf{w})&:=\Omega(\mathbf{w})+ \lambda\dot{R}(\mathbf{w},\mathbf{v},D)
\end{align}
\normalsize
\subsection{Self paced subclass learning}
Inspired by the self-paced learning framework of \cite{Kumar10}, we pose this problem as the following optimization,\footnote{For the simplicity of the notation, we assume samples are ordered by their labels thus in the set $D$ there are $N_p$ positive ($y_i=1$) samples indexed by $i=1:N_p$ and then $N_n$ negative ($y_i=-1$) samples indexed by $i=N_p+1:N_p+N_n$.}
\small
\begin{align}
\label{eq_selftuned}
\dot{R}(\mathbf{w},\mathbf{v},D)&=\sum_{i=1}^{N_p}v_i(l_{i}^{\mathbf{w}}-1/K_p)+
\sum_{i=1}^{N_n}v_{i+N_p}(l_{i+N_p}^{\mathbf{w}}-1/K_n)\\
&\mathbf{v}=\{v_1,v_2,...,v_{N}\}, v_i\in\{0,1\}\nonumber
\end{align}
\normalsize
where $\mathbf{v}$ is a binary indicator vector that denotes which samples are considered in the learning of $\mathbf{w}$ parameters. Proportional to $\lambda$, $K_p$ and $K_n$ are scalars which control how predictable a sample should be to get in the subset. The larger the $K$ the lower the loss of a sample should be in order to influence the learning of $\mathbf{w}$. The inclusion of more low loss samples ($l^{w}_{i} \leq 1/K$) is encouraged since it will decrease the total risk. Thus, $K$ provides a trade-off between the inclusion of samples in learning and the amount of irregularity in $\mathbf{w}$ needed for that. This is in line with our intended trade-off of finding the largest subset while maintaining a level of regularity. Large $K$ is a tighter constraint on the inclusion of the samples based on their closeness to the rest of the data. \\
This formulation is similar to Kumar et. al self-paced learning \cite{Kumar10} where $K_p=K_n$. \cite{Kumar10} proposed this framework for the non-convex latent variable models and in order to help them to converge to better local optima. \\
Optimizing \ref{eq_selftuned} directly, involves integer programming for variable $\textbf{v}$. However, for a convex choice of $l$ and $\Omega$ the relaxed version (i.e. $v_i \in \mathbb{R}, 0 \leq v_i \leq 1$) is a biconvex programming in $\mathbf{w}$ and $\mathbf{v}$ whose optimal value can be shown to be at $v_i \in \{0,1\}$. This means that when we fix either of the subset of variables $\mathbf{w}$ or $\mathbf{v}$ the problem becomes convex over the other subset. Biconvex problems are specially interesting since a simple alternating method has shown to converge to relatively good local optima efficiently \cite{Gorski07}.

\subsection{Ramp loss expert learning}
Finding the largest subset with high margin from negative samples can be seen as an outlier exclusion problem. That is, we want to detect those samples that are not coherent with (outliers w.r.t.) the largest majority of the data and exclude them from learning. This can be achieved by augmenting the loss function as follows,

\small
\begin{align}
\label{eq_ramp}
\dot{R}(\mathbf{w},D)&=\sum_{i=1}^{N_p}min(l_{i}^{\mathbf{w}}, S_p)+
\sum_{i=1}^{N_n}min(l_{i+N_p}^{\mathbf{w}}, S_n)
\end{align}
\normalsize

where the loss induced by each sample is capped at $S_p$ for positive samples and $S_n$ for negative samples. This means that those samples whose loss exceed the corresponding threshold will be discounted from optimization. One should note that, given specific model parameters $\mathbf{w}$, the gradient of high loss samples ($l^{w}_{i} > S$) is zero. Therefore, those samples does not influence the total risk by the change of $\mathbf{w}$ in any direction. It can be shown that the solution $\mathbf{w}^*$ of the above formulation is the same as if we have trained the model only on the samples with their $l^{w}_{i} <= S$. $S$ in combination with $\lambda$ controls the trade-off between regularization and inclusion of more samples. A large enough $S$ will include all samples and as we decrease it towards it will be more strict to include samples. \\
This formulation is a general case of ramp loss SVM \cite{Collobert06} where we have L2 regularization and hinge loss. \cite{Collobert06} proposed ramp loss SVM to decrease the number of support vectors. This was in order to address scalability issues associated with learning non-linear SVMs. \\
It should be noted that for a convex choice of regularization $\Omega$ and sample loss functions \ref{eq_ramp} can be reformulated as difference of convex functions,

\begin{figure*}[t]
  \mbox{}\\
  \centering
  \includegraphics[clip=true, trim=0cm 0cm 0cm 0cm, width=0.6\textwidth]{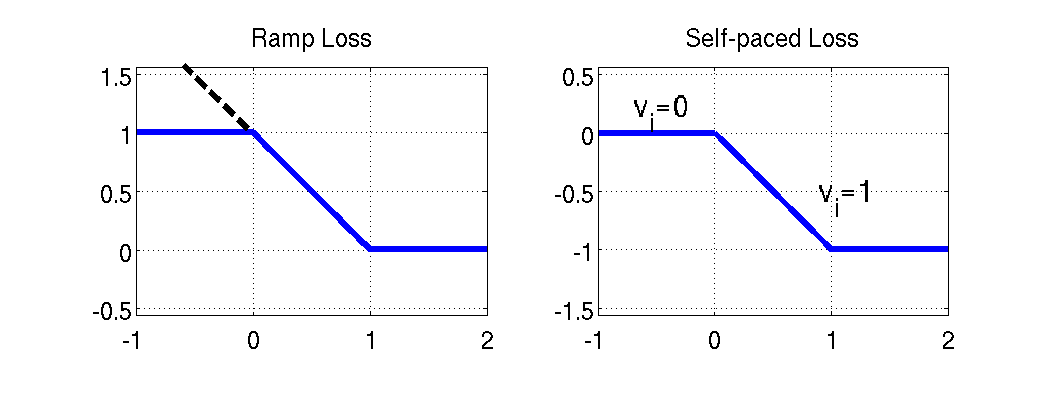}  
  \caption{The loss that sample $i$ induces with $S=1/K=1$ for the case of Ramp Loss (left) and Self-paced Loss (right). Dashed line shows the original hinge loss.}
  \label{fig:equivalence}
    
\end{figure*}

\tiny
\begin{align}
\label{eq_doc}
\dot{R}(\mathbf{w},D)&=\underbrace{\sum_{i=1}^{N_p}l_{i}^{\mathbf{w}}+
\sum_{i=1}^{N_n}l_{i+N_p}^{\mathbf{w}}}_{\text{convex}}-\underbrace{[\sum_{i=1}^{N_p}max(l_{i}^{\mathbf{w}}-S_p,0)+\sum_{i=1}^{N_n}max(l_{i+N_p}^{\mathbf{w}}-S_n, 0)]}_{\text{convex}}
\end{align}
\normalsize

This is true since $l$s are convex by choice, $S_p$ and $S_n$ are constants and convexity is preserved under max and summation operations. (\ref{eq_doc}) can be solved using concave convex procedure (CCCP) \cite{Yuille03}.
\subsection{Equivalence show case}
It can be shown that the solutions to these two minimizations are equivalent. Here we give an intuition. Consider the case that we only have a single parameter $S=S_p=S_n=1/K_p=1/Kn$. Furthermore, take hinge loss as $l_{i}^{\mathbf{w}}$. In self-paced equation (\ref{eq_selftuned}), for a given $\mathbf{w}$, we have the optimal values of indicator variables $v_i=1$ iff $l_{i}^{\mathbf{w}}<S$ and $v_i=0$ otherwise. That means, each sample $i$ induces a self-paced loss,

\small
\begin{align}
\label{eq_selfpaced_equivalence}
v_i(l_{i}^{\mathbf{w}}-S) = \left\{ 
  \begin{array}{l l}
    max(1-y_is_i, 0) - S & \quad \forall{i} \, l_{i}^{\mathbf{w}}<S\\
    0 & \quad \text{otherwise}
  \end{array} \right.
\end{align}
\normalsize
On the other hand each samples induced ramp loss is $min(max(1-y_is_i,0),S)$ by definition \cite{Collobert06}. With a closer look at these two one see that self-paced induced loss is a shifted version of ramp loss by $-S$ and thus their solution are equivalent (look at figure \ref{fig:equivalence}).

\subsection{Equivalence}

It can be shown that the solutions to these two minimization are equivalent. Let's assume the $\mathbf{w}^*_r$ is the solution to the minimization  (5). Without loss of generality, let's assume we have one parameter $S=S_p=S_n=1/K_p=1/Kn$. That gives,

\tiny
\begin{align}
\label{eq_rampOpt}
J^{*}_{ramp}=\Omega(\mathbf{w}^{*}_r)+\sum_{i=1}^{N}min(l_{i}^{\mathbf{w}^{*}_r}, S) \; \leq \; \Omega(w)+\sum_{i=1}^{N}min(l_{i}^{\mathbf{w}}, S) \qquad \forall{\mathbf{w}}
\end{align}
\normalsize
Now, let's assume that $\mathbb{N}^{\mathbf{w}}_i$ is the set of inlier samples that have an induced loss of $l^{\mathbf{w}}_{i} \leq S$ w.r.t. $\mathbf{w}$. And $\mathbb{N}^{\mathbf{w}}_o$ is the set of remaining samples that are outliers ($|\mathbb{N}^{\mathbf{w}}_i|+|\mathbb{N}^{\mathbf{w}}_o|=N$). Then from \ref{eq_rampOpt} we have,

\tiny
\begin{align}
\label{eq_rampOpt2}
&J^{*}_{ramp}=\Omega(\mathbf{w}^{*}_r)+\sum_{i\in\mathbb{N}^{\mathbf{w}^{*}_r}_i}l_{i}^{\mathbf{w}^{*}_r} + |\mathbb{N}^{\mathbf{w}^{*}_r}_o|S \; \leq \;  \Omega(\mathbf{w})+\sum_{i\in\mathbb{N}^{\mathbf{w}}_i}l_{i}^{\mathbf{w}}+ |\mathbb{N}^{\mathbf{w}}_o|S \qquad \forall{\mathbf{w}}
\end{align} 
\normalsize

On the other hand, let's assume $\mathbf{w}^*_s$ is one optimal solution to minimization (4). Optimal values of $v_i$ will be one for all inliers and 0 for outliers as otherwise it would not be an optimal point. We also have $S=1/K_p=1/K_n$. Thus, 

\tiny
\begin{align}
\label{eq_selfOpt2}
\nonumber  &J^{*}_{self}=\Omega(\mathbf{w}^{*}_s)+\sum_{i\in\mathbb{N}^{\mathbf{w}^{*}_s}_i}l_{i}^{\mathbf{w}^{*}_s} - |\mathbb{N}^{\mathbf{w}^{*}_s}_i|S \; \leq \;  \Omega(\mathbf{w})+\sum_{i\in\mathbb{N}^{\mathbf{w}}_i}l_{i}^{\mathbf{w}}-|\mathbb{N}^{\mathbf{w}}_i|S \qquad \forall{\mathbf{w}}\\
&\overset{|\mathbb{N}^{\mathbf{w}}_i|+|\mathbb{N}^{\mathbf{w}}_o|=N}{\implies} \Omega(\mathbf{w}^{*}_s)+\sum_{i\in\mathbb{N}^{\mathbf{w}^{*}_s}_i}l_{i}^{\mathbf{w}^{*}_s} + |\mathbb{N}^{\mathbf{w}^{*}_s}_o|S \; \leq \;  \Omega(\mathbf{w})+\sum_{i\in\mathbb{N}^{\mathbf{w}}_i}l_{i}^{\mathbf{w}}+ |\mathbb{N}^{\mathbf{w}}_o|S \qquad \forall{\mathbf{w}}
\end{align} 
\normalsize

Comparing (\ref{eq_rampOpt2}) to (\ref{eq_selfOpt2}) shows that $\mathbf{w}^{*}_s$ and $\mathbf{w}^{*}_r$ are solutions to both minimizations (4) and (5).

\subsection{Discussion}
We want the visual subclasses to be well separated from \textit{all} the negative samples. So we are interested in the special case of the above formulations when $S_n>>0$ for ramp loss formulation and negative indicator variables set to 1 ($v_i=1$) for self-tuned formulation. This will ensure that all the negative samples are included in the learning. Thus, we assume there is only one parameter $K_p$ and $S_p$ which we simple call $K$ and $S$ and in general \textit{inclusion parameter}.\\
A larger choice of inclusion parameter will be a tighter constraint on the positive samples which are contributing to the final solution. That means by choosing large enough or small enough $K$ or $S$ in the formulations above we span a spectrum where we have exemplar learner \cite{Malisiewicz11} at one extreme and a monolithic learner at the other extreme respectively. \\ 
The two methods use different alternate searches to generate upper bound formulations (linear tangent in case of CCCP for ramp loss difference of convex formulation and fixed indicator variable for self-paced formulation). However, it can be shown that the generated bounds are closely related. It can be easily shown that the slight difference is the tendency of the ramp loss approximated bound to keep outliers having low scores (remaining outlier) while self-paced formulation completely ignores the outliers in its approximated bound.

{\small
\bibliographystyle{ieee}
\bibliography{local}
}

\end{document}